\documentclass{article}
\PassOptionsToPackage{numbers,sort&compress}{natbib}
\usepackage[main, preprint]{neurips_2026}

\usepackage[utf8]{inputenc} 
\usepackage[T1]{fontenc}    
\usepackage[pagebackref]{hyperref} 
\usepackage{url}            
\usepackage{booktabs}       
\usepackage{amsfonts}       
\usepackage{amsmath}
\usepackage{multicol}
\usepackage{makecell}
\usepackage{multirow}
\usepackage{nicefrac}       
\usepackage{microtype}      
\usepackage{xcolor}         
\usepackage{graphicx}
\usepackage[table]{xcolor}
\usepackage{pifont}
\usepackage{wrapfig}
\usepackage{cleveref}
\newcommand{\yes}{\ding{51}}
\newcommand{\no}{\ding{55}} 
\newcommand{\ours}{StreamMAE}

\title{I Have a Stream: Making Self-Supervised \\ Learning Work on Continuous Video}

\author{%
Ivan Martinović\textsuperscript{1,2} \quad
Lukas Knobel\textsuperscript{2} \quad
Yuki M.~Asano\textsuperscript{2}
\\
\textsuperscript{1}Faculty of Electrical Engineering and Computing, University of Zagreb \\
\textsuperscript{2}Fundamental AI Lab, University of Technology Nuremberg
\\
\texttt{ivan.martinovic@fer.hr}
\quad
\texttt{\{lukas.knobel,yuki.asano\}@utn.de}
\\
\url{https://martinovicivan.github.io/StreamMAE}
}

\newcommand{\mstd}[2]{${#1}_{\pm #2}$}
\begin{document}
\newcommand{\wtp}{\texttt{WT++12h}}
\newcommand{\wtpp}{\texttt{WT++25h}}
\newcommand{\wtppp}{\texttt{WT++50h}}
\newcommand{\wtpppp}{\texttt{WT++95h}}
\definecolor{changes}{RGB}{0,0,0}

\maketitle

\begin{abstract}
Self-supervised learning draws inspiration from infant visual development, yet standard training pipelines bear little resemblance to it: images are independently sampled and globally shuffled across epochs. 
We study self-supervised learning from continuous video streams, where frames are consumed in temporal order using strict sliding-window batches, without global reshuffling or multi-epoch replay. To this end, we construct \texttt{WT++}, a 95-hour urban walking-tour video dataset for streaming pretraining. Combined with a comprehensive evaluation suite 
we find that contrastive and distillation-based methods struggle in this setting, while MAE is more robust but still falls short of standard i.i.d.~pretraining. We find that high \textit{inter}-batch similarity, caused by sliding-window consumption across consecutive batches, does not explain this gap. The main challenge is high \textit{intra}-batch similarity, where frames within each batch are near-duplicates. To mitigate this, we propose \ours{}, which preserves the core MAE reconstruction objective while adapting the input pipeline with stream-aware regularization and motion-biased crop selection. \ours{} outperforms streaming baselines, matches i.i.d.~MAE trained on the same video data, remains competitive with ImageNet-pretrained MAE, and scales positively as the pretraining stream grows from 12 to 95 hours.
\end{abstract}

\section{Introduction}
\label{sec:introduction}
Self-supervised learning (SSL) methods, such as MAE~\cite{he2022masked} and DINO~\cite{caron2021emerging,oquab2023dinov2}, have advanced visual representation learning. These models are trained on large image collections~\cite{jia2009imagenet,ridnik2021imagenet21k,oquab2023dinov2} that are shuffled, often revisited over multiple epochs, and sampled to form diverse batches. This approximates an independent-and-identically-distributed (i.i.d.) regime at the batch level: batches contain diverse, unrelated examples, supporting stable optimization and broad coverage. While SSL is often motivated by how infants and animals learn without explicit high-level supervision, such as language, this pretraining regime remains far from how visual experience arrives in practice.

Natural visual experience is inherently sequential. A camera, robot, or embodied agent observes the world as a temporally ordered stream, where consecutive frames are highly similar and visual content evolves slowly over time. This setting removes a central convenience of standard gradient-based SSL: the ability to globally shuffle data and construct diverse batches. We therefore ask: \emph{Can SSL work \underline{well} on continuous video streams and \underline{scale} with more video data when trained from random initialization, without global shuffling, multi-epoch training, or a long-term replay buffer?}

Learning from streaming video is appealing not only as a fundamental research question, but also for embodied agents and edge devices, where data naturally arrive sequentially and need not be stored as a fixed training dataset. However, this regime challenges standard gradient-based training: instead of diverse batches that approximate the data distribution, streaming batches come from short consecutive video segments, are highly redundant, and often contain only small visual changes. As a result, SSL methods that work well on large image collections may not be the best fit for continuous video.

Several recent works study learning from continuous video, but typically relax the strict from-scratch streaming setting: they focus on online prediction or adaptation with pretrained initialization~\cite{carreira2024learning}, address correlated updates without fully closing the from-scratch degradation~\cite{han2025learning}, or use replay buffers to restore batch diversity~\cite{yang2026memorystoryboard}. In contrast, we study SSL from scratch: models consume video in temporal order using sliding-window batches and are trained without long-term replay buffers.

To study this setting, we extend the released WalkingTours (WT) dataset~\cite{venkataramanan2024dora} into \texttt{WT++}, a 95-hour collection of walking-tour videos for streaming pretraining. We evaluate learned representations across several downstream vision tasks, covering both close-to-domain and out-of-domain benchmarks.

\begin{figure}[t!]
    \centering
    \includegraphics[width=1.0\linewidth]{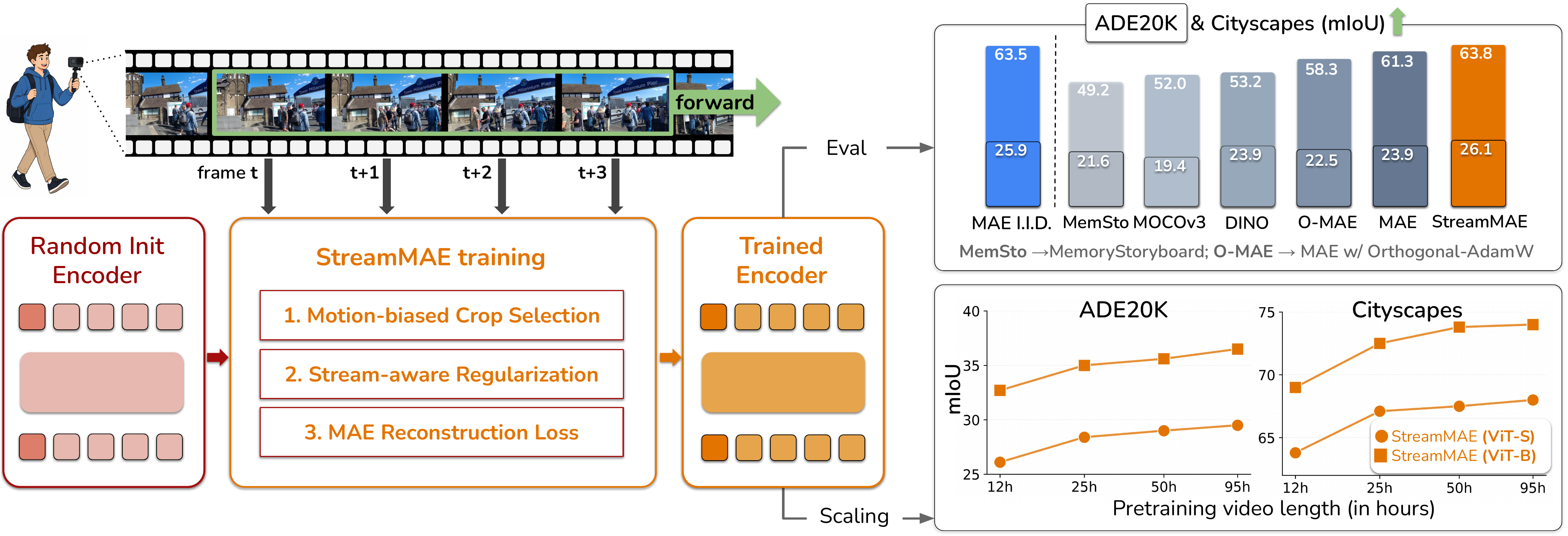}
\caption{
Left: \ours{} adapts MAE to continuous video streams with stream-aware regularization and crop selection. Right: \ours{} (ViT-S) improves dense performance over streaming SSL baselines (top) and improves with longer pretraining streams and a larger backbone (bottom).
}
\vspace{-1.3em}
    \label{fig:teaser}
\end{figure}

We first benchmark common SSL objectives under this streaming setup, including contrastive learning with MoCo v3~\cite{chen2021empirical}, self-distillation with DINO~\cite{caron2021emerging}, and masked reconstruction with MAE~\cite{he2022masked}. Under streaming pretraining, MoCo v3 and DINO lag behind MAE (see \Cref{fig:teaser}, right), making masked reconstruction the strongest starting point in our setting. However, streaming MAE still falls short of standard i.i.d.~MAE trained on the same visual data. We investigate this gap by separating two effects: \emph{inter-batch similarity}, i.e., similarity between consecutive batches caused by fixed-order sliding-window consumption, and \emph{intra-batch similarity}, i.e., similarity among examples within the same batch. To isolate these effects, we pre-shuffle ImageNet-1K~\cite{jia2009imagenet} once and then consume it as a fixed stream. MAE trained on this pre-shuffled stream matches standard i.i.d.~MAE, showing that inter-batch similarity alone does not explain the gap. The main difficulty instead comes from high intra-batch similarity in continuous video, where each batch contains many near-duplicate frames.

Motivated by this analysis, we propose \ours{}, a simple adaptation of MAE for streaming video. As summarized in \Cref{fig:teaser}, \ours{} preserves the MAE reconstruction objective while adapting the training pipeline to reduce the effect of temporally redundant batches. It combines stronger regularization, two-stage cropping, and motion-biased crop selection based on patch-level frame differences. \ours{} outperforms streaming baselines, narrows the gap to standard i.i.d.~MAE on the same video data, and scales with encoder capacity and pretraining duration (\Cref{fig:teaser}, bottom right). These results suggest that masked reconstruction, combined with stream-aware sampling and regularization, is a strong foundation for self-supervised learning from continuous video streams.

\section{Related Work}  \label{sec:related_work}

\textbf{Self-supervised image encoders.}
Self-supervised image encoders have become standard backbones for vision tasks, transferring well to classification, segmentation, depth estimation, 3D geometry, and other downstream tasks~\cite{yang2024depth,oquab2023dinov2,wang2023cut}. They are trained with a range of objectives. Contrastive methods, such as MoCo~\cite{he2020momentum,chen2020improved,chen2021empirical}, learn by contrasting different image views. Self-distillation methods, such as DINO and DINOv2~\cite{caron2021emerging,oquab2023dinov2}, align student and teacher representations. Masked modeling methods, such as MAE~\cite{he2022masked}, I-JEPA~\cite{assran2023ijepa}, and CAPI~\cite{darcet2025cluster}, learn from partially observed images by reconstructing pixels or predicting representations. Some works also train image encoders from video, often using temporal information through ordering or motion~\cite{salehi2023time,salehi2025mosic}, or by reconstructing future patches~\cite{siammae2023neurips}. Closely related to our data source, \citet{venkataramanan2024dora} and \citet{han2025uniquelivessharedworld} study SSL of image encoders from walking-tour videos. Despite their success, these methods are typically developed under i.i.d.-style pretraining on large image collections or offline video datasets, where samples can be shuffled and batches contain diverse examples. We instead study how standard image SSL objectives behave on continuous video streams with high intra-batch similarity.

\textbf{Learning from streaming video.} 
Video SSL has also been studied with explicitly temporal objectives, including temporal ordering~\cite{misra2016shuffle}, predictive coding~\cite{han2019video}, and masked video modeling~\cite{tong2022videomae,wang2023videomaev2}. However, most video SSL methods still assume offline access to videos, where clips can be sampled randomly, shuffled into batches, and revisited over multiple epochs. In contrast, we keep the encoder image-based and study a streaming setting in which frames are consumed in temporal order.

While several methods study continual SSL~\cite{fini2022self,hu2022how} or online learning from images~\cite{hayes2019memory,hayes2020remind}, only a few works consider self-supervised learning from continuous video streams directly. Prior methods often mitigate temporal correlation through replay or memory mechanisms, including infinite or minimum-redundancy replay buffers~\cite{zhuang2022well,purushwalkam2022challenges} and reservoir-based~\cite{vitter1985random} memory over temporally segmented videos~\cite{yang2026memorystoryboard}. 
\textcolor{changes}{Relatedly, \citet{mall2025cram} study supervised continual video classification using a rolling replay buffer of compressed video codes.}
\citet{carreira2024learning} study online learning from a single continuous stream, but focuses on online prediction and adaptation, with its strongest setups relying on ImageNet~\cite{jia2009imagenet} initialization.
\textcolor{changes}{\citet{wang2025testtime} use masked reconstruction to adapt a task-trained model during deployment, making their approach complementary to our pretraining setting.}
Closest to our work, \citet{han2025learning} target correlated gradient updates, but training from scratch remains substantially below settings initialized from pretrained models. In contrast, we study from-scratch representation learning from ordered video streams and adapt MAE through streaming-aware regularization and crop selection, without relying on long-term replay buffers.

\section{Learning from a Streaming Video}
\label{sec:streaming_learning_analysis}
\subsection{Streaming Sliding-Window Batches}

\textcolor{changes}{
We study a streaming learning setting in which a model is trained on a continuous video stream, with each batch formed from a sliding window that advances through the stream in fixed temporal order.
}
Let the raw video be:
\begin{equation}
\mathbf{V} = (\mathbf{x}_1, \mathbf{x}_2, \ldots, \mathbf{x}_N),
\qquad
\mathbf{V} \in \mathbb{R}^{N \times H \times W \times 3},
\end{equation}
where each frame $\mathbf{x}_i$ is an RGB image. Given a batch size $B$ and stride $s$, we form batches as sliding windows over the stream. With training steps indexed from $t=0$, two consecutive batches are:
\begin{equation}
\mathbf{X}^{(t)} = (\mathbf{x}_{1 + ts}, \ldots, \mathbf{x}_{B + ts}),
\qquad
\mathbf{X}^{(t+1)} = (\mathbf{x}_{1 + (t+1)s}, \ldots, \mathbf{x}_{B + (t+1)s}).
\end{equation}
Thus, when $s < B$, consecutive batches overlap by $B-s$ frames.

\begin{figure}[h!]
\vspace{-1em}
    \centering
    \includegraphics[width=0.7\linewidth]{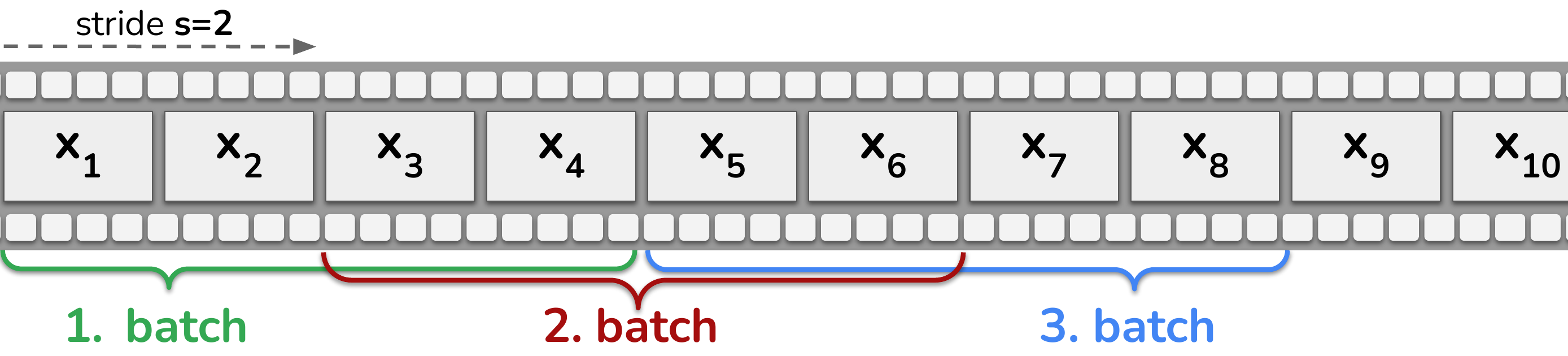}
    \caption{
    \textbf{Streaming pretraining setup.} Frames are observed in temporal order, and each batch is formed as a sliding window over the stream. Here: $B=4$ and stride $s=2$, but typically $s \ll B$.
    }
    \label{fig:streaming_setup}
    \vspace{-1.4em}
\end{figure}

We assume access to a batch of examples at each training step, since learning from a single frame at a time would impose an overly restrictive setting. This setup differs from standard i.i.d.~training, where samples are shuffled and batches are formed without preserving temporal structure. Figure~\ref{fig:streaming_setup} illustrates the described streaming setup for $s=2$ and $B=4$.

\subsection{\texttt{WT++} Dataset} \label{subsec:wtpp_dataset_main}
We train models on urban-scene walking-tour videos, following~\citet{venkataramanan2024dora} and the released WalkingTours (WT) dataset. WT consists of long, single-shot videos recorded while a person walks through different cities. These videos contain diverse objects, lighting conditions, viewpoints, and scene transitions, making them a natural testbed for studying learning from video streams.

The WT dataset contains approximately 13 hours of video, which limits larger-scale streaming experiments. We therefore extend it and construct \texttt{WT++}, a 95-hour dataset comprising 58 public walking-tour videos. \texttt{WT++} includes all videos from the original WT dataset~\cite{venkataramanan2024dora}, except for the Wildlife safari video.
Details are provided in Appendix~\ref{sec:wtppdataset}.

By default, models are trained on a single 12-hour video, \texttt{WT++London} (i.e.,~\texttt{WT++12h}). For larger-scale experiments, we concatenate multiple walking-tour videos into one ordered stream and train with the same streaming protocol. This approximates a long continuous stream, since publicly available single-shot walking-tour videos rarely span tens of hours. The stream remains temporally ordered within each video with discontinuities at video boundaries.

\begin{figure}[t]
    \centering
    \includegraphics[width=0.9\linewidth]{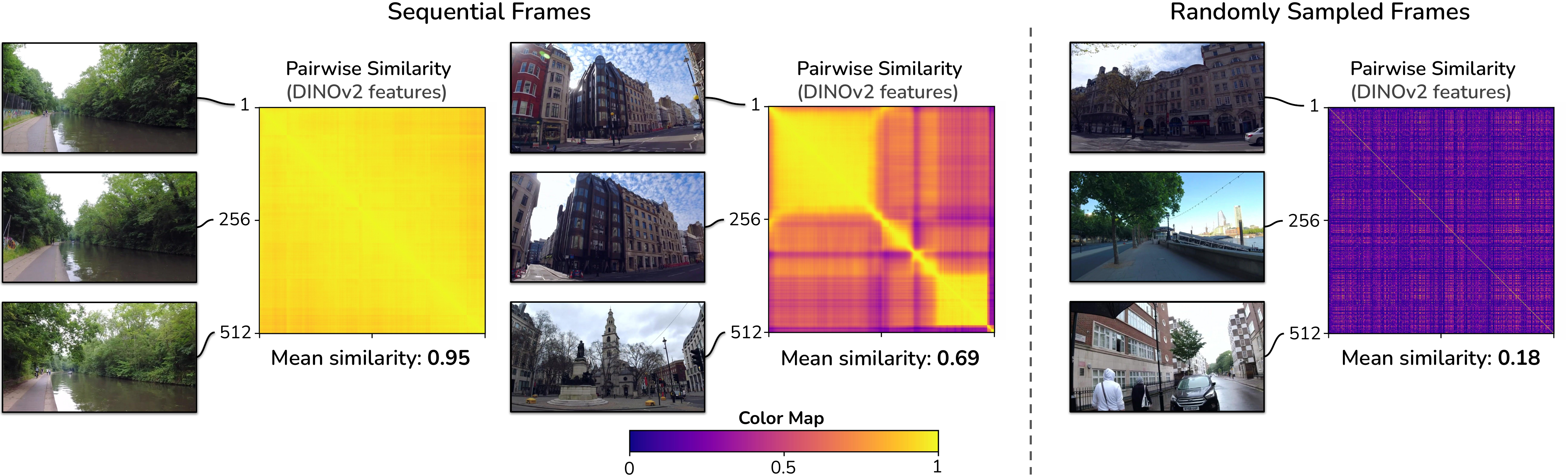}
    \caption{
Pairwise DINOv2-L~\cite{oquab2023dinov2} feature similarities for 512 frames from \texttt{WT++London} show that consecutive frames at \texttt{15 FPS} can be highly similar (left) or contain structured scene changes (block diagonal, middle), whereas randomly sampled frames are markedly more diverse (right).
}
\vspace{-1.4em}
    \label{fig:analyzing_of_frames_with_dv2_features}
\end{figure}

To illustrate the temporal structure of the data, we analyze frame similarity in \texttt{WT++London} using DINOv2~\cite{oquab2023dinov2} features. We compare pairwise cosine similarities for 512 consecutive frames sampled from a local stream window against 512 frames sampled uniformly at random from the same video.
As shown in Figure~\ref{fig:analyzing_of_frames_with_dv2_features}, consecutive 15~\texttt{FPS} frames can be highly similar, whereas randomly sampled frames are substantially more diverse. This illustrates the distributional gap between streaming and shuffled batches, and motivates methods that can learn from temporally local, correlated data.

\vspace{-0.8em}
\section{Towards \ours{}}
\label{sec:streammae}
Our goal is to design a self-supervised method that can learn from continuous video streams. As suggested by \Cref{fig:teaser}, MAE is more robust under streaming pretraining than MoCo v3~\cite{chen2021empirical} and DINO~\cite{caron2021emerging}. This is consistent with the nature of their objectives: masked reconstruction operates on each example independently, avoiding the need to contrast examples within similar batches~\cite{chen2021empirical} or to cluster visual concepts across a diverse pretraining corpus~\cite{caron2021emerging} - both of which become problematic when the data stream is temporally correlated. We therefore build on MAE and use standard i.i.d.~sampled pretraining on the same visual data as a reference throughout this work.

However, streaming MAE still falls short of this i.i.d.~reference (\textit{cf}.~\Cref{fig:teaser}), and the gap becomes substantial when scaling to ViT-B/16, underperforming by nearly 10 mIoU points on Cityscapes (\textit{cf}.~\Cref{tab:tab_scaling_the_encoder_size}). To understand this gap, we identify two key departures from standard i.i.d.~pretraining. The first is \emph{inter-batch similarity}: in streaming training, the fixed-order sliding-window consumption means that consecutive batches share most of their examples, leading to highly correlated gradient updates across successive optimization steps. The second is \emph{intra-batch similarity}: because each batch is drawn from a local temporal window of the video stream, the examples within a single batch can be near-duplicate frames depicting the same scene with only minor visual variations.

\begin{wraptable}{r}{0.4\linewidth}
\vspace{-1.5em}
\centering
\caption{Batch similarity statistics.}
\label{tab:batch_similarity_statistics}
\small
\begin{tabular}{lccc}
\toprule
& \multirow{2}{*}{\makecell[c]{i.i.d.\\IN-1K}} 
& \multicolumn{2}{c}{Streaming} \\
\cmidrule(lr){3-4}
Metric 
& 
& IN-1K 
& \wtp{} \\
\midrule
$\mu_{\mathrm{intra}}$ & 0.004 & 0.004 & 0.665 \\
$\mu_{\mathrm{inter}}$ & 0.325 & 0.989 & 1.000 \\
\bottomrule
\end{tabular}
\vspace{-1.5em}
\end{wraptable}
To disentangle these two factors, we construct a controlled experiment. We pre-shuffle ImageNet-1K once and treat the resulting sequence as a fixed stream, forming batches with a sliding window of stride $s=8$. This retains the fixed-order sliding-window consumption of the streaming protocol, and therefore high inter-batch similarity, while removing the near-duplicate frames that characterize video-stream batches, yielding low intra-batch similarity. For each setting (IN-1K-i.i.d., IN-1K pre-shuffled+streaming, and our \wtp{}), we draw consecutive batches of frames and measure inter-batch and intra-batch cosine similarity using features from a pretrained DINOv2~\cite{oquab2023dinov2} model: $\mu_{\mathrm{inter}}$, $\mu_{\mathrm{intra}}$ (details in the Appendix~\ref{sec:appendix_sims}).

As shown in Table~\ref{tab:batch_similarity_statistics}, the three settings span a spectrum. The IN-1K i.i.d.\ setting exhibits low similarity on both metrics, as expected from random sampling over a diverse dataset. The \wtp{} stream yields substantially higher values for both: $\mu_{\mathrm{intra}}=0.665$ reflects high visual similarity within each temporal window, and $\mu_{\mathrm{inter}}=1.000$ reflects nearly complete overlap between successive sliding-window batches. Crucially, the IN-1K pre-shuffled streaming setting occupies an informative middle ground: it nearly matches the \wtp{} stream in inter-batch similarity ($\mu_{\mathrm{inter}}=0.989$) due to the same sliding-window mechanics, but retains the low intra-batch similarity of i.i.d.\ training ($\mu_{\mathrm{intra}}=0.004$) because the underlying images are diverse. This decoupled setting allows us to isolate the effect of each factor, which we investigate next.

\subsection{Preliminary Analysis}
\begin{wraptable}{r}{0.5\linewidth}
    \vspace{-1.6em}
    \centering
    \footnotesize
    \caption{
Fixed-order sliding-window training on diverse data. We pre-shuffle ImageNet-1K once and consume it as a fixed stream with stride $s=8$. The pre-shuffled stream matches standard i.i.d.~MAE, suggesting that inter-batch similarity alone does not explain the streaming gap.}
    \vspace{0.2em}
    \label{tab:imagenet_streaming}
    \resizebox{\linewidth}{!}{%
        \setlength{\tabcolsep}{2.2pt}
\begin{tabular}{llccc}
    \toprule
    \multirow{2}{*}{Backbone} &
    \multirow{2}{*}{Training regime} &
    \multirow{2}{*}{\makecell[c]{IN-1K \\ Acc@1~$\uparrow$}} &
    \multirow{2}{*}{\makecell[c]{CS \\ mIoU~$\uparrow$}} &
    \multirow{2}{*}{\makecell[c]{ADE20K \\ mIoU~$\uparrow$}} \\
    & & & & \\
    \midrule
    \multirow{2}{*}{ViT-S}
        & Standard i.i.d.      & 77.4 & 64.0 & 26.9 \\
        & Pre-shuffled stream  & 77.4 & 63.6 & 27.2 \\
    \midrule
    \multirow{2}{*}{ViT-B}
        & Standard i.i.d.      & 81.5 & 72.3 & 35.9 \\
        & Pre-shuffled stream  & 81.6 & 73.6 & 36.5 \\
    \bottomrule
\end{tabular}

    }
    \vspace{-1em}
\end{wraptable}%
\textbf{Does inter-batch similarity explain the gap?} 
Given the gap between streaming MAE and standard i.i.d.~MAE, %
we utilize the pre-shuffled IN-1K stream to analyze whether inter-batch similarity is leading to degraded representations. As shown in \Cref{tab:imagenet_streaming}, a pre-shuffled IN-1K stream matches standard i.i.d.~MAE across several downstream benchmarks. 
\textcolor{changes}{A controlled comparison on \wtp{} shows the same trend
(Appendix~\ref{sec:appendix_wt_similarity}).} These results suggest that \emph{for MAE, inter-batch similarity alone is not harmful when batches remain visually diverse. The main challenge is high intra-batch similarity induced by near-duplicate frames in a continuous video.
}

\begin{figure}[b!]
\color{changes}
\vspace{-1em}
    \centering
    \includegraphics[width=0.82\linewidth]{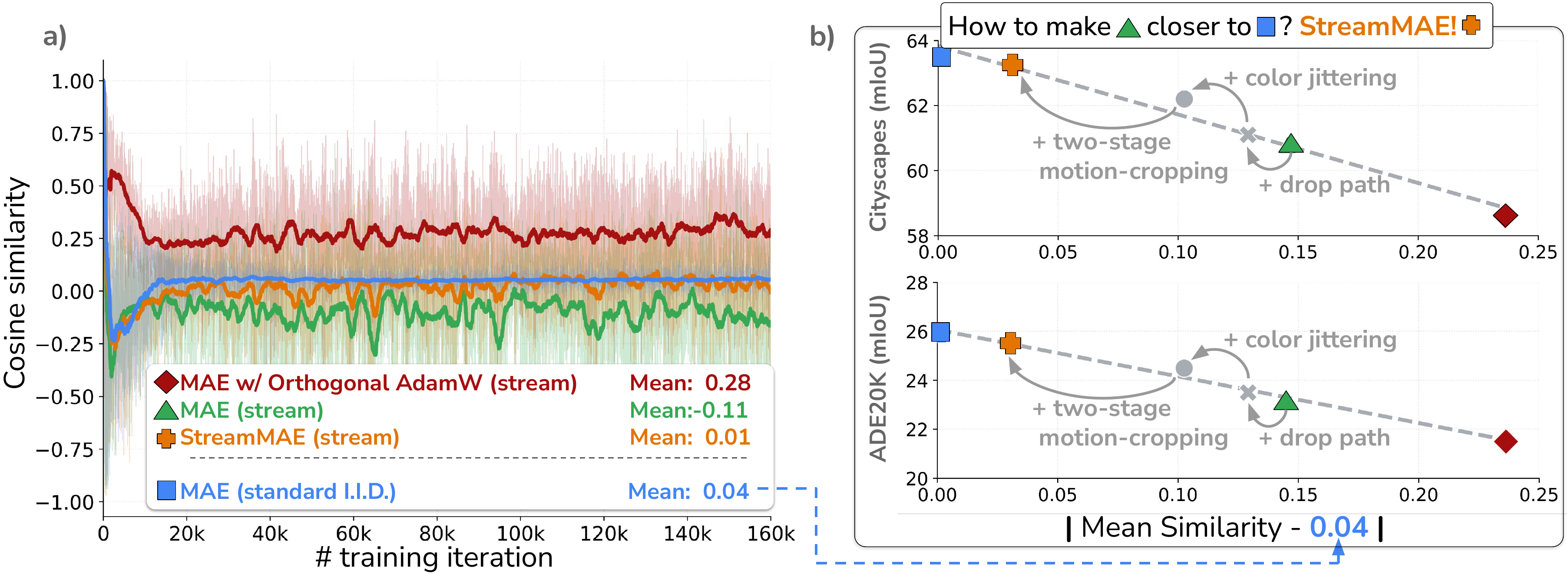}
    \caption{
(a) Running average of consecutive batches gradient similarity, measured on last-block MLP parameters every 10 steps.
(b) Dense downstream performance versus distance to standard i.i.d.~MAE in mean gradient similarity.
Methods closer to the i.i.d.~MAE reference perform better.}
    \label{fig:gradient_similarity_figure}
    \vspace{-1.4em}
\end{figure}

\noindent\textbf{How does high intra-batch similarity affect optimization?}
We next analyze whether the gap between standard i.i.d.~and streaming pretraining is reflected in the optimization dynamics of MAE-based methods. Following \citet{han2025learning}, we measure the cosine similarity between gradients from consecutive batches, using the MLP parameters in the last transformer block. \Cref{fig:gradient_similarity_figure}(a) shows the running average (with window 100) of this similarity throughout training, together with the temporal mean for each method. Interestingly, while \citet{han2025learning} emphasize highly positive gradient correlations in streaming training of DoRA~\cite{venkataramanan2024dora} (a method closer to DINO), we find that vanilla streaming MAE can produce weakly negative consecutive-gradient similarity, whereas MAE with Orthogonal-AdamW~\cite{han2025learning} yields substantially positive similarity. Standard i.i.d.~MAE, used as our reference, has more stable near-zero similarity. Here we use $B=512$ for all methods. 

\Cref{fig:gradient_similarity_figure}(b) relates dense downstream performance to the distance between each method's mean consecutive-gradient similarity and that of standard i.i.d.~MAE. Within the MAE family, methods closer to the i.i.d.~MAE reference achieve stronger dense performance. This suggests that aligning streaming optimization dynamics with standard i.i.d.~MAE can be a useful design principle. Motivated by this observation, \ours{} keeps the MAE objective unchanged, but adapts the training pipeline through stronger regularization, two-stage cropping, and motion-biased crop selection. As shown in \Cref{fig:gradient_similarity_figure}, these changes bring the gradient behavior closer to i.i.d.~MAE and improve downstream performance. We describe the components next.

\subsection{Regularization under High Intra-Batch Similarity}

Streaming video often produces batches with high intra-batch similarity: consecutive frames can contain near-duplicate views of the same scene (\textit{cf.}~\Cref{fig:analyzing_of_frames_with_dv2_features}). In this regime, standard MAE may rely on short-term regularities, repeatedly reconstructing similar content with limited variation in appearance or structure. We therefore strengthen the MAE training pipeline with three lightweight mechanisms: color jitter, increased drop path rate, and \text{DataDrop}.

\noindent\textbf{Color jitter.}
Color jitter increases appearance diversity by perturbing low-level statistics such as color, brightness, and contrast. This makes reconstruction less dependent on stable appearance cues that persist across neighboring frames.

\noindent\textbf{Drop path.}
Increasing the drop path rate~\cite{huang2016droppath} regularizes the encoder by varying the effective network depth across updates. This helps reduce overfitting to recurring local patterns in batches with high intra-batch similarity.

\noindent\textbf{\text{DataDrop}.}
In streaming pretraining, each batch corresponds to a sliding window from the ordered stream (\Cref{fig:streaming_setup}). When this window contains many highly similar frames, backpropagating through all examples can overemphasize near-duplicate content. We therefore use \text{DataDrop}: for each batch, we sample a binary mask over examples and compute the loss only on the retained subset. Dropped examples are not forwarded through the model and do not contribute to the current update. Training then proceeds to the next window according to the stream stride $s$. By default, we drop $75\%$ of examples in each batch, reducing the effective batch size for backpropagation by a factor of four.

\subsection{Stream-Aware Cropping}
\noindent\textbf{Two-stage cropping.}
MAE applies random resized cropping directly to the input frame. In streaming video, however, adjacent frames are often globally similar, so independently sampled crops from neighboring frames can still show near-duplicate content. We therefore use a two-stage cropping procedure that exposes the crop location as a separate design choice. First, we sample a fixed-size crop from the frame, defining a local spatial region. We then apply the standard MAE random resized crop within this region. This preserves the original MAE augmentation pipeline while making the sampled view depend on an explicit first-stage region selection.

This formulation separates \emph{where} to crop from \emph{how} to augment the final training view. A simple version samples the first-stage region uniformly at random. We next make this selection stream-aware by biasing it toward regions with recent visual change.

\begin{figure}[h!]
    \centering
    \includegraphics[width=1.0\linewidth]{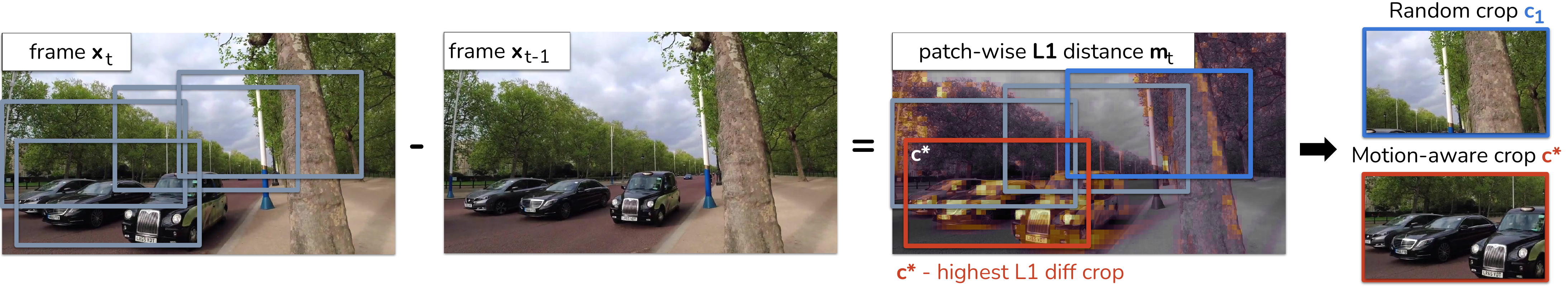}
    \caption{
    Motion-biased crop selection computes patch-level differences between consecutive frames $\mathbf{x}_t$ and $\mathbf{x}_{t-1}$, samples $K$ crop candidates $\mathcal{C}$, and selects the crop $c_t^\star$ with the largest average $\ell_1$ score. 
    }
    \label{fig:example_of_selected_crops_randomly_vs_motionbiased}
    \vspace{-1.4em}
\end{figure}

\noindent\textbf{Motion-biased crop selection.}
Let $\mathbf{x}_t \in \mathbb{R}^{H \times W \times 3}$ denote the current frame and $\mathbf{x}_{t-1}$ the previous frame. We compute a per-pixel frame-difference map:
\begin{equation}
\mathbf{D}_t(h,w)
=
\left\|
\mathbf{x}_t(h,w,:) - \mathbf{x}_{t-1}(h,w,:)
\right\|_1,
\end{equation}
where $(h,w)$ indexes spatial locations and the $\ell_1$ norm is taken over RGB channels.
This provides a lightweight estimate of local temporal change without requiring optical flow~\cite{rlt2024neurips}. We aggregate $\mathbf{D}_t$ at the same patch granularity as the ViT encoder. Let $\Omega_p$ denote the set of pixel locations contained in patch $p$. The motion score of patch $p$ is:
\begin{equation}
m_t(p)
=
\frac{1}{|\Omega_p|}
\sum_{(h,w) \in \Omega_p}
\mathbf{D}_t(h,w).
\end{equation}
The resulting patch-level map $\mathbf{m}_t$ highlights regions of the current frame that differ most from the preceding frame (\emph{cf}.~\Cref{fig:example_of_selected_crops_randomly_vs_motionbiased}). We use this map to select the first-stage crop. Given $K$ first-stage candidate crops $\mathcal{C} = \{c_1,\ldots,c_K\}$, let $\mathcal{P}(c)$ denote the set of ViT patches covered by crop $c$. We select the candidate with the largest average patch-level motion:
\begin{equation}
c_t^\star
=
\arg\max_{c \in \mathcal{C}}
\frac{1}{|\mathcal{P}(c)|}
\sum_{p \in \mathcal{P}(c)}
m_t(p).
\end{equation}
We then apply the standard MAE random resized crop within $c_t^\star$. Thus, motion-biased crop selection preserves the MAE objective and final augmentation pipeline while biasing the first-stage region toward parts of the video that change over time. The case $K=1$ reduces to the random two-stage cropping procedure described above. In practice, we apply motion-biased selection with probability $0.5$ and otherwise fall back to standard random resized cropping. This stochastic choice avoids an overly deterministic focus on the same high-motion regions in  video segments.

\section{Experiments}   \label{sec:experiments}

\noindent\textbf{Training details.}
We use ViT-Small (ViT-S/16)~\cite{dosovitsky2021vit} as our main architecture and report ViT-B/16 results for \ours{} when scaling encoder size. All SSL methods are implemented in solo-learn~\cite{sololearn2022turrisi} and trained from scratch. We use AdamW~\cite{loshchilov2018decoupled} with $\beta_1=0.9$, $\beta_2=0.95$, and base learning rates $8 \times 10^{-5}$ for ViT-S and $4 \times 10^{-5}$ for ViT-B. The learning rate is scaled linearly with the effective batch size, $B_{\mathrm{eff}}/256$. Streaming runs use a constant learning rate with linear warm-up, consistent with the goal of continuing pretraining as new videos arrive.

Our default pretraining stream is \wtp{}, a 12-hour London walking-tour video from \texttt{WT++}. For larger-scale pretraining, we sequentially append videos to form \wtpp{}, \wtppp{}, and \wtpppp{}, corresponding to approximately 25, 50, and 95 hours of data. Each stream extends the preceding one by appending new videos while retaining all earlier videos in the same order (\emph{cf}.~ Appendix~\ref{sec:wtppdataset}). The original videos are recorded at 60~FPS and processed at $1280 \times 720$. We temporally subsample them by a factor of $k=16$, yielding 3.75~FPS. Streaming batches use stride $s=1$, except for \wtpppp{} experiments where we use $s=2$ to reduce training time. We form local stream windows of size $B=2048$ and apply \text{DataDrop} with drop probability $0.75$, yielding effective batch size $B_{\mathrm{eff}}=512$. 

For MAE-based methods, we use the standard masking ratio $0.75$ with random uniform masking~\cite{he2022masked}. We compare this choice with two alternatives in Appendix~\ref{sec:appendix_masking_strategy}. The default \ours{} uses color jitter ($p=0.5$), a linear drop-path schedule reaching $0.25$ in the final block, 
  DataDrop, two-stage cropping, and motion-biased crop selection with $K=4$ candidate crops. Further implementation details, including crop selection and compute requirements, are provided in Appendix~\ref{sec:appendix_training_details} and ~\ref{sec:appendix_compute}.

\noindent\textbf{Evaluation.}
\textcolor{changes}{Following established MAE evaluation protocols~\cite{he2022masked} and recent recommendations for semantic segmentation~\cite{kerssies2024benchmark}, we use end-to-end fine-tuning as our primary evaluation protocol. We additionally report attentive probing for classification in \Cref{tab:attentive_probing} to evaluate the frozen representations}. For classification, we use ImageNet-1K~\cite{jia2009imagenet}. For dense prediction, we use close-to-domain outdoor benchmarks, Cityscapes~\cite{Cordts2016Cityscapes} and KITTI~\cite{geiger2012kitti}, and out-of-domain benchmarks, ADE20K~\cite{zhou2017ade} and NYU-Depth-v2~\cite{silberman2012nyu}. We initialize the encoder from the pretrained checkpoint and fine-tune it with a task-specific head: a linear head for segmentation and DPT~\cite{ranftl2021vision} for depth estimation. We report top-1 val accuracy for classification, mIoU for segmentation, and RMSE for depth estimation. For depth estimation, we report \mstd{\text{mean}}{\text{std}} over two seeds. More details are provided in Appendix~\ref{sec:appendix_evaluation_details}.

\subsection{Benchmarking Self-Supervised Methods under Streaming Pretraining}

\begin{table}[h!]
\centering
\caption{
\textbf{Benchmarking SSL methods under streaming pretraining (ViT-S).}
We compare \wtp{} streaming methods to random initialization, i.i.d.~MAE on \wtp{}, and an iteration-matched ImageNet-1K i.i.d.~MAE reference. $^{\dagger}$ marks reproduced streaming-tailored methods. 
}  
    \footnotesize
    \label{tab:benchmarking_streaming_methods}
    
\setlength{\tabcolsep}{5.0pt}
\begin{tabular}{lcccccc}
        \toprule
        \multirow{2}{*}{\makecell[c]{Pretraining \\ Method}}  & \multirow{2}{*}{\makecell[c]{Pretraining \\ Data}} &
        \multirow{2}{*}{\makecell[c]{IN-1K~\cite{jia2009imagenet} \\ Acc @ 1 $\uparrow$}}  &
        \multirow{2}{*}{\makecell[c]{City~\cite{Cordts2016Cityscapes} \\ mIoU $\uparrow$}} & 
        \multirow{2}{*}{\makecell[c]{ADE~\cite{zhou2017ade} \\ mIoU $\uparrow$}} & 
        \multirow{2}{*}{\makecell[c]{NYUv2~\cite{silberman2012nyu} \\ RMSE $\downarrow$}} & 
        \multirow{2}{*}{\makecell[c]{KITTI~\cite{geiger2012kitti} \\ RMSE $\downarrow$}} \\
        & & & & & &  \\
        \midrule
        Random Init  & -- & 71.9 & 47.7 & 16.7 & \mstd{0.877}{0.013} & \mstd{6.108}{0.028} \\
        \midrule
        \rowcolor{gray!7} \multicolumn{7}{c}{Standard I.I.D.~Setup} \\
        \rowcolor{gray!7} MAE -- standard IID      &  \wtp{}          & 77.0 & 63.5 & 25.9 & \mstd{0.701}{0.004} & \mstd{4.070}{0.006}  \\
        \rowcolor{gray!7} MAE -- standard IID  & ImageNet-1K      & 77.4 & 64.0 & 26.9 & \mstd{0.656}{0.004} & \mstd{3.854}{0.008}  \\
        \midrule
         \multicolumn{7}{c}{Streaming Setup} \\
        MOCO-v3~\cite{chen2021empirical}    & \wtp{} & 68.7 & 52.0 & 19.4 & \mstd{0.760}{0.000} & \mstd{4.882}{0.054} \\
        DINO~\cite{caron2021emerging}       & \wtp{} & 72.7 & 53.2 & 23.9 & \mstd{0.734}{0.003} & \mstd{4.454}{0.034}  \\
        MAE~\cite{he2022masked}             & \wtp{} & 77.1 & 61.3 & 23.9 & \mstd{0.742}{0.002} & \mstd{4.211}{0.137}  \\
        
        Orthogonal-MAE~\cite{han2025learning}$^\dagger$ & \wtp{} & 75.9 & 58.3 & 22.5 & \mstd{0.749}{0.004} & \mstd{4.209}{0.036}  \\
         MemoryStoryboard~\cite{yang2026memorystoryboard}$^\dagger$ & \wtp{} & 71.0 & 49.2 & 21.6 & \mstd{0.792}{0.001} & \mstd{4.647}{0.038}  \\
        \rowcolor{blue!3} \text{\ours{} (ours)} & \wtp{} & \textbf{77.5} & \textbf{63.8} & \textbf{26.1} & \textbf{\mstd{\mathbf{0.694}}{\mathbf{0.003}}} & \mstd{\mathbf{3.976}}{\mathbf{0.012}}  \\
        \midrule
        \rowcolor{blue!8} \ours{} (ours) 
        &\wtpppp & \textbf{78.4} & \textbf{68.0} & \textbf{29.5} & \mstd{\mathbf{0.646}}{\mathbf{0.003}} & \mstd{\mathbf{3.765}}{\mathbf{0.047}} \\
        \bottomrule
    \end{tabular}
    \vspace{-0.8em}
\end{table}

We first benchmark representative SSL methods under our streaming protocol and evaluate representations with full fine-tuning on downstream tasks. \textcolor{changes}{Details of baseline tuning and adaptations to temporal redundancy are provided in Appendices~\ref{sec:memory_storyboard_results} and~\ref{sec:appendix_baseline_tuning}.}
As shown in Table~\ref{tab:benchmarking_streaming_methods}, MAE is a stronger streaming baseline than MoCo v3~\cite{chen2021empirical} and DINO~\cite{caron2021emerging}. MoCo v3 even falls below random initialization on ImageNet-1K, consistent with contrastive learning being sensitive to high intra-batch similarity where near-duplicate frames can act as false negatives. DINO is more stable, but still trails MAE on most downstream tasks. Streaming-tailored baselines do not close the gap. MAE with Orthogonal-AdamW~\cite{han2025learning} underperforms standard streaming MAE, and our reproduction of MemoryStoryboard~\cite{yang2026memorystoryboard} with ViT-S remains below MAE despite using a replay buffer (\textit{cf.}~Appendix~\ref{sec:memory_storyboard_results}). In contrast, \ours{} improves over all streaming baselines and is comparable to standard IID MAE trained on the same \wtp{} data. Scaling \ours{} to \wtpppp{} further improves transfer, especially on dense prediction. We also include an ImageNet-1K IID MAE reference trained for the same number of iterations as the \wtp{} streaming runs. 

\begin{figure}[h!]
    \centering
    \includegraphics[width=0.7\linewidth]{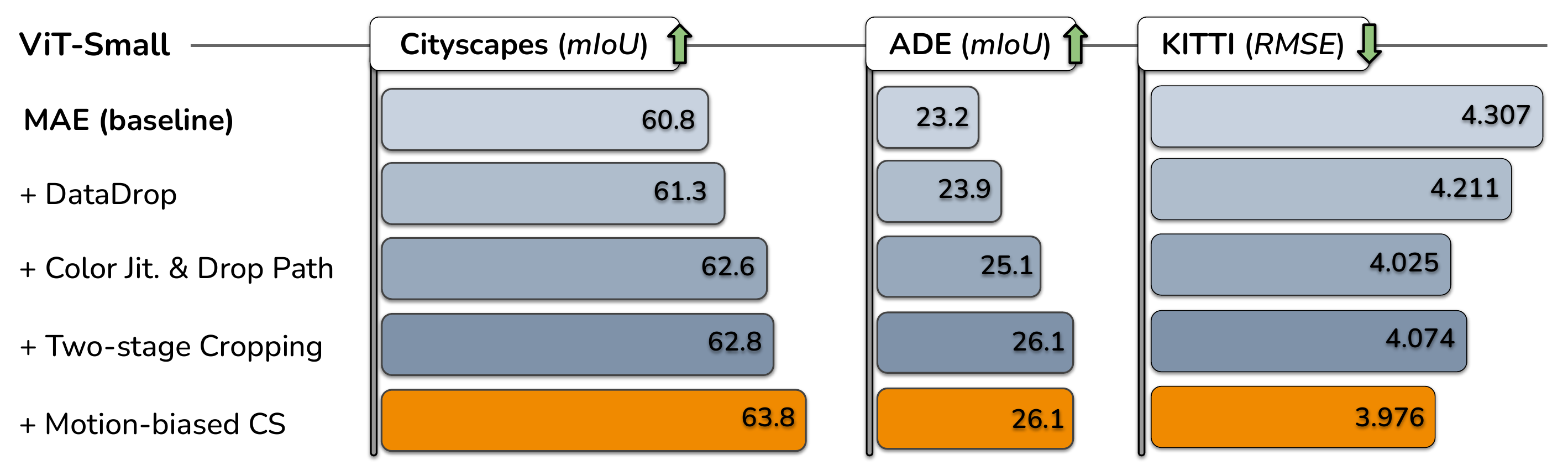}
    \caption{
\textbf{Cumulative ablation study with ViT-S/16 pretrained on \wtp{}}. Each row adds one component to the previous setting. \text{DataDrop} gives modest gains, while stronger regularization and crop selection account for most of the dense-transfer improvement. CS denotes crop selection. 
}
\label{fig:ablation_study_fig}
\vspace{-1.4em}
\end{figure}

\noindent\textbf{Ablation study.}
We perform cumulative ablations using a ViT-S/16 backbone pretrained on \wtp{}. As shown in Figure~\ref{fig:ablation_study_fig}, the proposed design choices progressively improve dense transfer. \text{DataDrop} yields modest gains, while stronger regularization and crop selection account for most of the improvement. Motion-biased crop selection gives the largest gains on close-to-domain outdoor benchmarks, Cityscapes~\cite{Cordts2016Cityscapes} and KITTI~\cite{geiger2012kitti}. Later experiments show that \text{DataDrop} becomes negligible at larger scale, whereas regularization and crop selection remain the main components. The MAE baseline in this ablation uses $B=512$, which gives stronger performance in our setup (\textit{cf.}~Appendix~\ref{subsec:onbaselines_and_datadrop}). 

\subsection{Scaling and Analysis}
We next analyze the scaling behavior of \ours{} by varying encoder capacity and pretraining duration, and examine whether its design choices remain beneficial at larger scale.

\noindent\textbf{Scaling encoder capacity.}
We scale the encoder to ViT-B/16 while keeping the pretraining stream fixed to \wtp{}. As shown in \Cref{tab:tab_scaling_the_encoder_size}, increasing model capacity alone does not close the gap between streaming baseline and standard i.i.d. MAE, particularly on dense prediction. Orthogonal-MAE provides modest gains over streaming MAE, but remains below \ours{}. In contrast, \ours{} scales effectively: it nearly matches standard i.i.d. MAE on ImageNet-1K and ADE20K, surpasses it on Cityscapes, and substantially improves depth estimation over streaming MAE. The ImageNet-1K i.i.d. MAE reference is trained for the same number of iterations as the \wtp{} runs.
\begin{table}[h!]
    \centering
    \caption{
\textbf{Increasing encoder size}. \ours{} (ViT-B) improves over both \ours{} (ViT-S) and the MAE baseline, while closing most of the gap to standard i.i.d.~MAE. $^{\dagger}$ marks our reproduction.
}
    \setlength{\tabcolsep}{4.5pt}
    \footnotesize
    \label{tab:tab_scaling_the_encoder_size}

\setlength{\tabcolsep}{3pt}
\begin{tabular}{lccccccc}
        \toprule
        \multirow{2}{*}{\makecell[c]{Pretraining \\ Method}}  &
        \multirow{2}{*}{\makecell[c]{Encoder}}  &
        \multirow{2}{*}{\makecell[c]{Pretraining \\ Data}} &
        \multirow{2}{*}{\makecell[c]{IN-1K\\ Acc @ 1 $\uparrow$}}  &
        \multirow{2}{*}{\makecell[c]{Cityscapes\\ mIoU $\uparrow$}} & 
        \multirow{2}{*}{\makecell[c]{ADE20K\\ mIoU $\uparrow$}} & 
        \multirow{2}{*}{\makecell[c]{NYUv2\\ RMSE $\downarrow$}} & 
        \multirow{2}{*}{\makecell[c]{KITTI\\ RMSE $\downarrow$}} \\
        & & & & & & &  \\
        \midrule
        Random Init & ViT-B & -- & \text{78.5} & \text{51.0} & \text{17.5} & \mstd{0.881}{0.002} & \mstd{5.818}{0.137} \\
        \midrule
        \rowcolor{gray!7} \multicolumn{8}{c}{Standard I.I.D.~Setup} \\
        \rowcolor{gray!7}MAE -- standard IID  & ViT-B&\wtp          & \text{81.2} & \text{67.8} & \text{32.9} & \mstd{0.675}{0.004} & \mstd{3.855}{0.048}  \\
        \rowcolor{gray!7}MAE -- standard IID  & ViT-B&ImageNet-1K & \text{81.5} & \text{72.3} & \text{35.9} & \mstd{0.588}{0.003} & \mstd{3.771}{0.039}  \\
        \midrule
         \multicolumn{8}{c}{Streaming Setup} \\
        \rowcolor{blue!3} \text{\ours{} (ours)} & ViT-S &\wtp{} & \text{77.5} & \text{63.8} & \text{26.1} & \textbf{\mstd{{0.694}}{{0.003}}} & \mstd{{3.976}}{{0.012}}  \\
        MAE~\cite{he2022masked} & ViT-B&\wtp{} & 80.0 & 58.2 & 26.2 & \mstd{0.753}{0.005} & \mstd{4.543}{0.046}  \\
        Orthogonal-MAE~\cite{han2025learning}$^\dagger$ & ViT-B & \wtp{} & \text{80.0} & \text{60.6} & \text{27.3} & \mstd{0.750}{0.002} & \mstd{4.289}{0.011}  \\
        \rowcolor{blue!3} \text{\ours{} (ours)} & ViT-B & \wtp{} & \textbf{81.1} & \textbf{69.0} & \textbf{32.7} & \mstd{\mathbf{0.655}}{\mathbf{0.002}} & \mstd{\mathbf{3.797}}{\mathbf{0.006}}  \\
        \midrule
        \rowcolor{blue!8} \text{\ours{} (ours)} &ViT-B  &  \wtpppp & \textbf{82.0} & \textbf{74.0} & \textbf{36.5} &
             \mstd{\mathbf{0.584}}{\mathbf{0.002}} & 
             \mstd{\mathbf{3.496}}{\mathbf{0.009}} \\
        \bottomrule
    \end{tabular}
\end{table}

\noindent\textbf{Scaling pretraining duration.}
We next scale the pretraining stream by sequentially appending additional walking-tour videos to \wtp{}. As shown in \Cref{tab:scaling_data_and_models}, performance generally improves with longer streams for both ViT-S/16 and ViT-B/16. \textcolor{changes}{We additionally compare the \wtpppp{} runs against standard i.i.d.~MAE pretrained on ImageNet-1K under the same longer budget of $650\mathrm{k}$ iterations, as reported in Appendix~\Cref{tab:increasing_in1k_iterations}. With ViT-S/16, \ours{} matches this longer reference on IN-1K and exceeds it on both segmentation benchmarks. With ViT-B/16, it remains competitive on IN-1K and Cityscapes but trails on ADE20K. Overall, these results show that \ours{} benefits from additional streaming data while remaining close to the longer-budget, iteration-matched i.i.d.~MAE reference on most benchmarks.}

We further analyze the role of \text{DataDrop} under this scaling regime in~\Cref{tab:removing_fifo_when_scaling}. While \text{DataDrop} provides modest gains in the default ViT-S/16 setting, its effect becomes negligible for ViT-B/16 when scaling pretraining to \wtppp{} or \wtpppp{}. This suggests that \text{DataDrop} is useful at smaller scale (see also \Cref{fig:ablation_study_fig}), but is not the main driver of performance once model capacity and stream duration are increased. Here, the no-\text{DataDrop} setting uses $B=512$, corresponding to roughly 136 seconds of video at 3.75~FPS. Additional batch-size sensitivity results are provided in Appendix~\ref{sec:sensitivity_to_batch_size}.

\begin{table}[t]
    \centering
    \footnotesize
    \setlength{\tabcolsep}{2.5pt}

    \begin{minipage}[t]{0.56\linewidth}
        \centering
        \caption{
        \textbf{Scaling pretraining video stream duration.}
        }
        \label{tab:scaling_data_and_models}
        \resizebox{\linewidth}{!}{%
            \setlength{\tabcolsep}{1.2pt}
\begin{tabular}{l@{\quad}lccccc}
        \toprule
        &

        \multirow{1}{*}{\texttt{WT} duration}  &
        \multirow{1}{*}{\makecell[c]{IN-1K~$\uparrow$}} &
        \multirow{1}{*}{\makecell[c]{CS~$\uparrow$}} &
        \multirow{1}{*}{\makecell[c]{ADE~$\uparrow$}} &
        \multirow{1}{*}{\makecell[c]{NYUv2~$\downarrow$}} &
        \multirow{1}{*}{\makecell[c]{KITTI~$\downarrow$}}\\
        \midrule
        \multirow{4}{*}{\rotatebox{90}{ViT-S}}
        &\wtp & \text{77.5} & \text{63.8} & \text{26.1} & \mstd{0.694}{0.003} & \mstd{3.976}{0.012} \\
        
        &\wtpp & \text{78.1} & \text{67.1} & \text{28.4} & \mstd{0.659}{0.003} & \mstd{\mathbf{3.747}}{\mathbf{0.048}} \\
        
        &\wtppp & \text{78.2} & \text{67.5} & \text{29.0} & \mstd{0.656}{0.001} & \mstd{3.895}{0.041} \\
        &\wtpppp & \textbf{78.4} & \textbf{68.0} & \textbf{29.5} & \mstd{\mathbf{0.646}}{\mathbf{0.003}} & \mstd{3.765}{0.047} \\
        \midrule
        \multirow{4}{*}{\rotatebox{90}{ViT-B}}
        &     \wtp & 81.1 & 69.0 & 32.7 & 
             \mstd{0.655}{0.002} & 
             \mstd{3.797}{0.006} \\
        &     \wtpp & 81.6 & 72.5 & 35.0 &
             \mstd{0.617}{0.003} & 
             \mstd{3.619}{0.009} \\
        &     \wtppp & 81.9 & 73.8 & 35.6 &
             \mstd{0.600}{0.003} & 
             \mstd{3.620}{0.023} \\

        &    \wtpppp & \textbf{82.0} & \textbf{74.0} & \textbf{36.5} &
             \mstd{\mathbf{0.584}}{\mathbf{0.002}} & 
             \mstd{\mathbf{3.496}}{\mathbf{0.009}} \\
    
            \bottomrule
    \end{tabular}
        }
    \end{minipage}
    \hfill
    \begin{minipage}[t]{0.43\linewidth}
        \centering
        \caption{
\textbf{Effect of DataDrop at scale.} Its contribution is negligible for ViT-B on longer pretraining videos.
}
        \vspace{-3.2pt}
        \label{tab:removing_fifo_when_scaling}
        
        \resizebox{\linewidth}{!}{%
            \setlength{\tabcolsep}{2.2pt}
\begin{tabular}{lccccc}
        \toprule
         &
        \multirow{2}{*}{\texttt{WT} duration}  &
        \multirow{2}{*}{\makecell[c]{\texttt{DataDrop}}}  &
        \multirow{2}{*}{\makecell[c]{IN-1K \\ Acc@1~$\uparrow$}} &
        \multirow{2}{*}{\makecell[c]{CS \\ (mIoU)~$\uparrow$}} &
        \multirow{2}{*}{\makecell[c]{ADE \\ (mIoU)~$\uparrow$}} \\
        & & & & & \\        
        \midrule
        
        \multirow{2}{*}{\rotatebox{90}{ViT-S}}
            & \wtppp    & \yes & 78.2 & 67.5 & 29.0 \\
            & \wtppp    & \no  & 78.3 & 66.0 & 29.4 \\
            \midrule
            
        \multirow{4}{*}{\rotatebox{90}{ViT-B}}
            & \wtppp    & \yes & 81.9 & 73.8 & 35.6 \\
            & \wtppp    & \no  & 81.7 & 73.6 & 36.4 \\
            \cmidrule{2-6}
            & \wtpppp    & \yes & 82.0 & 74.0 & 36.5 \\
            & \wtpppp    & \no  & 81.9  & 74.3   & 36.3 \\
        \bottomrule
    \end{tabular}
        }
    \end{minipage}
    \vspace{-1.4em}
\end{table}

\begin{wraptable}{r}{0.48\linewidth}
    \vspace{-1.6em}
    \centering
    \footnotesize
    \caption{
    \textbf{Checkpoint averaging} on the same ViT-B \wtpppp{} trajectory improves dense transfer with negligible change in IN-1K accuracy.
    }
    \label{tab:checkpoint_averaging}
    \resizebox{\linewidth}{!}{%
        \setlength{\tabcolsep}{1.2pt}
\begin{tabular}{lcccc}
        \toprule
        \multirow{2}{*}{Method} &
        \multirow{2}{*}{\texttt{WT} duration}  &
        \multirow{2}{*}{\makecell[c]{IN-1K \\ Acc@1~$\uparrow$}} &
        \multirow{2}{*}{\makecell[c]{CS \\ (mIoU)~$\uparrow$}} &
        \multirow{2}{*}{\makecell[c]{ADE \\ (mIoU)~$\uparrow$}} \\
        & & & & \\        
        \midrule
        \multirow{1}{*}{\ours{}}
            & \wtpppp       & \textbf{82.0} & 74.0 & 36.5 \\
            \cmidrule{2-5}
          Avg. \{40,60,80,100\}\%  & \wtpppp          & 81.9 & \textbf{75.5} & 37.0 \\
          Avg. \{60,100\}\%        & \wtpppp         &81.9 & 75.3 & \textbf{37.3} \\
            
        \bottomrule
    \end{tabular}

    }
    \vspace{-1.0em}
\end{wraptable}
\textbf{Checkpoint averaging for long streaming runs.}
Inspired by continual learning setups~\cite{marczak2024magmax,roth2024a}, we additionally examine whether averaging checkpoints from a long streaming run can improve the final representation. For ViT-B/16 pretrained on \wtpppp{}, we average weights from checkpoints saved at different stages of the same training trajectory. Percentages denote the fraction of training completed at each checkpoint. As shown in Table~\ref{tab:checkpoint_averaging}, this simple procedure improves dense prediction performance while leaving IN-1K accuracy essentially unchanged. These results suggest that checkpoints from different stages of pretraining may retain complementary information.

\noindent\textbf{More updates versus new visual content.}
We next ask whether scaling gains come from more optimization steps or from new visual content. Table~\ref{tab:increasing_fps} shows that increasing the effective number of updates by using higher \texttt{FPS} does not improve transfer and can hurt dense prediction.
\begin{table}[b!]
    \vspace{-0.8em}
    \centering
    \footnotesize
    \begin{minipage}[t]{0.48\linewidth}
        \centering
        \vspace{0pt}
        \caption{
        \textbf{Effect of temporal subsampling on \wtp{} with ViT-B.} Denser sampling increases the effective number of updates, but does not improve performance.
        }
        \label{tab:increasing_fps}
        \setlength{\tabcolsep}{2.2pt}
\begin{tabular}{lccc}
        \toprule
         \multirow{2}{*}{\makecell[c]{Subsampling \\ Factor ($k$)}} & 
         \multirow{2}{*}{\makecell[c]{IN-1K \\ Acc@1~$\uparrow$}} &
        \multirow{2}{*}{\makecell[c]{CS \\ (mIoU)~$\uparrow$}} &
        \multirow{2}{*}{\makecell[c]{ADE \\ (mIoU)~$\uparrow$}} \\
         & & & \\
        \midrule
         $k=4$ \texttt{(15 FPS)}           &\text{81.0}   & \text{65.2} & \text{30.1}  \\
         $k=8$ \texttt{(7.5 FPS)}          &\text{81.0}   & \text{66.8} & \text{31.4}  \\
        \rowcolor{blue!5}   $k=16$ \texttt{(3.75 FPS)}       &\textbf{81.1}   & \textbf{69.0} & \textbf{32.7} \\
        \bottomrule
    \end{tabular}

    \end{minipage}
    \hfill
    \begin{minipage}[t]{0.5\linewidth}
        \centering
        \vspace{0pt}
        \caption{
        \textbf{Effect of repeating the data.}
        Additional passes over \wtp{} improve performance, but new videos yield larger gains.
        }
        \label{tab:repeating_vs_new}
        \setlength{\tabcolsep}{6pt}
\begin{tabular}{lccc}
        \toprule
        \multirow{1}{*}{{\texttt{WT++} data}} &  \multirow{1}{*}{\makecell[c]{IN-1K}} &
        \multirow{1}{*}{{CS}} &
        \multirow{1}{*}{{ADE}} \\
        \midrule
             \wtp          &\text{81.1} & \text{69.0} & \text{32.7} \\
             \cmidrule{2-4}
             \wtp$\times$2   &81.3 & 70.3 & 33.3 \\
             \wtpp           &81.6 & 72.5 & 35.0 \\
             \cmidrule{2-4}
             \wtp$\times$4   &81.5 & 71.7 & 33.9  \\
             \wtppp          &\textbf{81.9} & \textbf{73.8} & \textbf{35.6} \\
        \bottomrule
    \end{tabular}

    \end{minipage}
\end{table}
Table~\ref{tab:repeating_vs_new} shows that \ours{} can still benefit from additional epochs over the same stream, indicating that it does not require extreme data diversity to improve. However, streams that include new videos perform better than repeated epochs over \wtp{}, suggesting that newly introduced scenes provide additional gains. These results suggest that longer-stream improvements are not explained by more updates alone. \ours{} benefits from both continued optimization and increased visual diversity.

\begin{wraptable}{r}{0.48\linewidth}
    \vspace{-1.6em}
    \begingroup
    \color{changes}
    \centering
    \caption{
    ImageNet-1K attentive probing evaluation after pretraining on \wtp{}.
    }
    \label{tab:attentive_probing}
    \scriptsize
    \setlength{\tabcolsep}{2.5pt}
        \begin{tabular}{@{}lcccc@{}}
        \toprule
        \multirow{2}{*}{Method}
        & \multicolumn{2}{c}{ViT-S/16}
        & \multicolumn{2}{c}{ViT-B/16} \\
        \cmidrule(lr){2-3}
        \cmidrule(lr){4-5}
        & Acc@1~$\uparrow$ & Acc@5~$\uparrow$
        & Acc@1~$\uparrow$ & Acc@5~$\uparrow$ \\
        \midrule
        MAE -- standard i.i.d.
        & 40.0 & 63.0
        & 46.4 & 69.0 \\
        MAE (streaming)
        & 33.4 & 55.8
        & 37.3 & 59.6 \\
        \ours{}
        & \textbf{40.2} & \textbf{63.6}
        & \textbf{47.1} & \textbf{69.9} \\
        \bottomrule
    \end{tabular}
    \vspace{-1.6em}
    \endgroup
\end{wraptable}

\begingroup
\color{changes}
\paragraph{Attentive probing.}
We evaluate frozen representations on IN-1K by training only a single-query cross-attention pooling layer followed by a linear classifier. We follow the probing hyperparameters of~\citet{psomas2026attention} and the augmentations of AIMv2~\cite{fini2025multimodal}. As shown in \Cref{tab:attentive_probing}, \ours{} improves over streaming MAE by $6.8$ and $9.8$ Acc@1 points for ViT-S/16 and ViT-B/16, respectively, while matching same-data i.i.d. MAE.

\subsection{Generalization Across Pretraining Domains}

To examine whether \ours{} extends beyond walking-tour video, we additionally pretrain ViT-B/16 on three diverse streams: HD-EPIC~\cite{perret2025hdepic}, an indoor egocentric kitchen stream, CROWD~\cite{alam2026global}, a front-facing urban dashcam stream, and KrishnaCAM~\cite{singh2016krishnacam}, a longitudinal egocentric daily-life stream. For each pretraining dataset, we compare streaming MAE, \ours{}, and standard i.i.d. MAE using the same frames and number of pretraining iterations. 

\begin{table}[h]
    \color{changes}
    \centering
    \caption{
    \textbf{Generalization across pretraining domains using ViT-B/16}.}
    \label{tab:domain_generalization}
    \footnotesize
    \setlength{\tabcolsep}{2.5pt}
    \renewcommand{\arraystretch}{1.05}
    \begin{tabular}{@{}llcccc@{}}
    \toprule
    \multirow{2}{*}{\makecell[c]{Pretraining \\ Data}} &
    \multirow{2}{*}{Method} &
    \multirow{2}{*}{\makecell[c]{CS \\ (mIoU)~$\uparrow$}} &
    \multirow{2}{*}{\makecell[c]{ADE \\ (mIoU)~$\uparrow$}} &
    \multirow{2}{*}{\makecell[c]{NYUv2 \\ (RMSE)~$\downarrow$}} &
    \multirow{2}{*}{\makecell[c]{KITTI \\ (RMSE)~$\downarrow$}} \\
    & & & & & \\
    \midrule
    \multirow{3}{*}{HD-EPIC~\cite{perret2025hdepic}}
    & MAE -- standard i.i.d. 
    & 65.2 & 30.7
    & \mstd{0.681}{0.001}
    & \mstd{4.184}{0.081} \\
    & MAE (streaming)
    & 63.2 & 29.3
    & \mstd{0.711}{0.000}
    & \mstd{4.206}{0.003} \\
     \rowcolor{blue!3} 
    & 
    \ours{} (ours)
    & \textbf{65.6} & \textbf{32.0}
    & \mstd{\mathbf{0.652}}{\mathbf{0.001}}
    & \mstd{\mathbf{4.030}}{\mathbf{0.049}} \\
    \midrule
    \multirow{3}{*}{CROWD~\cite{alam2026global}}
    & MAE -- standard i.i.d.
    & 68.1 & 32.3
    & \mstd{0.681}{0.000}
    & \mstd{3.830}{0.026} \\
    & MAE (streaming)
    & 65.7 & 29.1
    & \mstd{0.704}{0.004}
    & \mstd{4.287}{0.147} \\
    \rowcolor{blue!3} & \ours{} (ours)
    & \textbf{71.5} & \textbf{32.6}
    & \mstd{\mathbf{0.644}}{\mathbf{0.001}}
    & \mstd{\mathbf{3.756}}{\mathbf{0.046}} \\
    \midrule
    \multirow{3}{*}{KrishnaCAM~\cite{singh2016krishnacam}}
    & MAE -- standard i.i.d.
    & 71.7 & \textbf{34.6}
    & \mstd{\mathbf{0.616}}{\mathbf{0.004}}
    & \mstd{3.596}{0.003} \\
    & MAE (streaming)
    & 67.9 & 32.9
    & \mstd{0.663}{0.009}
    & \mstd{3.799}{0.010} \\
    \rowcolor{blue!3}&
    \ours{} (ours)
    & \textbf{72.2} & 34.5
    & \mstd{0.621}{0.001}
    & \mstd{\mathbf{3.583}}{\mathbf{0.012}} \\
    \bottomrule
\end{tabular}
    \vspace{-0.8em}
\end{table}

As shown in \Cref{tab:domain_generalization}, \ours{} consistently improves over streaming MAE across all four downstream tasks and all three domains, with gains of $1.6$--$5.8$ mIoU on segmentation and reductions of $0.042$--$0.531$ RMSE on depth estimation. On HD-EPIC and CROWD, \ours{} matches or exceeds same-data i.i.d. MAE on every downstream task. On KrishnaCAM, it trails i.i.d. MAE by $0.1$ mIoU on ADE20K and $0.005$ RMSE on NYUv2, while outperforming it on Cityscapes and KITTI. Details of preprocessing and stream construction for all three datasets are provided in Appendix~\ref{sec:appendix_additional_datasets_construction}.

\paragraph{Robustness to irregular camera motion.}
We further ablate motion-biased crop selection on KrishnaCAM~\cite{singh2016krishnacam}, whose head-mounted viewpoint exhibits irregular camera motion across indoor and outdoor scenes. Relative to \ours{} without motion-biased crop selection, the full method improves ADE20K and Cityscapes by $0.6$ and $0.7$ mIoU, respectively, and reduces KITTI RMSE by $0.063$, while leaving NYUv2 unchanged (full results are provided in Appendix \Cref{tab:krishnacam_motion_crops}). Motion-biased crop selection therefore remains beneficial beyond the walking-tour domain.
\endgroup
\section{Conclusion}
\label{sec:conclusion}
We studied self-supervised learning from continuous video, where models are trained from scratch on temporally ordered frames without global reshuffling or long-term replay buffers. We showed that contrastive~\cite{chen2021empirical} and self-distillation~\cite{caron2021emerging} methods struggle in this setting, while masked reconstruction provides a more robust starting point. Our analysis suggests that the main challenge for MAE is not inter-batch similarity from fixed-order sliding-window consumption, but high intra-batch similarity caused by near-duplicate frames. Based on this insight, we introduced \ours{}, which keeps the MAE objective while adapting the training pipeline to
mitigate the effects of high intra-batch similarity
and focus learning on more informative regions of the stream. \ours{} outperforms streaming baselines, is competitive with standard i.i.d.~MAE, and scales to longer video streams.

\clearpage
\begin{ack}
We thank Tengda Han for inspiring discussions and Ryousuke Yamada, Ivan Grubišić, Josip Šarić, Valentinos Pariza, and Ivan Sabolić for their feedback on the manuscript.
Ivan Martinović was supported by the Croatian Science Foundation through grants DOK-NPOO-2023-10-2288 and MOBDOK-2023-4880.
The authors gratefully acknowledge the scientific support and HPC resources provided by the Erlangen National High Performance Computing Center (NHR@FAU) of the Friedrich-Alexander-Universität Erlangen-Nürnberg (FAU) under the BayernKI project \texttt{v115be}. BayernKI funding is provided by Bavarian state authorities.
\end{ack}

\bibliographystyle{plainnat}
\bibliography{main}

@String(CVPR  = {IEEE Conf. Comput. Vis. Pattern Recog.})

@String(ICCV  = {Int. Conf. Comput. Vis.})

@String(ECCV  = {Eur. Conf. Comput. Vis.})

@String(NeurIPS = {Adv. Neural Inform. Process. Syst.})

@String(ICML  = {Int. Conf. Mach. Learn.})

@String(ICLR  = {Int. Conf. Learn. Represent.})

@String(CVPRW = {IEEE Conf. Comput. Vis. Pattern Recog. Worksh.})

@String(JMLR  = {J. Mach. Learn. Res.})

@String(TMLR  = {Trans. Mach. Learn Res.})

@String(CVPR  = {CVPR})

@String(ICCV  = {ICCV})

@String(ECCV  = {ECCV})

@String(NeurIPS = {NeurIPS})

@String(ICML  = {ICML})

@String(ICLR  = {ICLR})

@String(CVPRW = {CVPRW})

@String(JMLR  = {JMLR})

@String(TMLR  = {TMLR})

@inproceedings{caron2021emerging,
  title={Emerging properties in self-supervised vision transformers},
  author={Caron, Mathilde and Touvron, Hugo and Misra, Ishan and J{\'e}gou, Herv{\'e} and Mairal, Julien and Bojanowski, Piotr and Joulin, Armand},
  booktitle={ICCV},
  year={2021}
}

@inproceedings{he2022masked,
  title={Masked autoencoders are scalable vision learners},
  author={He, Kaiming and Chen, Xinlei and Xie, Saining and Li, Yanghao and Doll{\'a}r, Piotr and Girshick, Ross},
  booktitle={CVPR},
  pages={16000--16009},
  year={2022}
}

@article{
oquab2023dinov2,
title={{DINO}v2: Learning Robust Visual Features without Supervision},
author={Maxime Oquab and Timoth{\'e}e Darcet and Th{\'e}o Moutakanni and Huy V. Vo and Marc Szafraniec and Vasil Khalidov and Pierre Fernandez and Daniel HAZIZA and Francisco Massa and Alaaeldin El-Nouby and Mido Assran and Nicolas Ballas and Wojciech Galuba and Russell Howes and Po-Yao Huang and Shang-Wen Li and Ishan Misra and Michael Rabbat and Vasu Sharma and Gabriel Synnaeve and Hu Xu and Herve Jegou and Julien Mairal and Patrick Labatut and Armand Joulin and Piotr Bojanowski},
journal={TMLR},
issn={2835-8856},
year={2024},
}

@inproceedings{carreira2024learning,
  title={Learning from one continuous video stream},
  author={Carreira, Jo{\~a}o and King, Michael and Patraucean, Viorica and Gokay, Dilara and Ionescu, Catalin and Yang, Yi and Zoran, Daniel and Heyward, Joseph and Doersch, Carl and Aytar, Yusuf and others},
  booktitle=CVPR,
  year={2024}
}

@inproceedings{han2025learning,
  title={Learning from streaming video with orthogonal gradients},
  author={Han, Tengda and Gokay, Dilara and Heyward, Joseph and Zhang, Chuhan and Zoran, Daniel and Patraucean, Viorica and Carreira, Joao and Damen, Dima and Zisserman, Andrew},
  booktitle=CVPR,
  year={2025}
}

@InProceedings{yang2026memorystoryboard,
  title = 	 {{Memory Storyboard: Leveraging Temporal Segmentation for Streaming Self-Supervised Learning from Egocentric Videos}},
  author =       {Yang, Yanlai and Ren, Mengye},
  booktitle = 	 {Proceedings of The 4th Conference on Lifelong Learning Agents},
  year = 	 {2025},
}

@inproceedings{
venkataramanan2024dora,
title={{Is ImageNet worth 1 video? Learning strong image encoders from 1 long unlabelled video}},
author={Shashanka Venkataramanan and Mamshad Nayeem Rizve and Joao Carreira and Yuki M Asano and Yannis Avrithis},
booktitle={ICLR},
year={2024},
}

@article{sololearn2022turrisi,
  author  = {Victor Guilherme Turrisi da Costa and Enrico Fini and Moin Nabi and Nicu Sebe and Elisa Ricci},
  title   = {{solo-learn: A Library of Self-supervised Methods for Visual Representation Learning}},
  journal = JMLR,
  year    = {2022}
}

@article{yang2025pixio,
  title={{In Pursuit of Pixel Supervision for Visual Pre-training}},
  author={Yang, Lihe and Li, Shang-Wen and Li, Yang and Lei, Xinjie and Wang, Dong and Mohamed, Abdelrahman and Zhao, Hengshuang and Xu, Hu},
  journal={arXiv:2512.15715},
  year={2025}
}

@article{lowe2026bootleg,
      title={{Self-Distillation of Hidden Layers for Self-Supervised Representation Learning}}, 
      author={Scott C. Lowe and Anthony Fuller and Sageev Oore and Evan Shelhamer and Graham W. Taylor},
      journal={arXiv:2603.15553},
      year={2026},
}

@inproceedings{assran2023ijepa,
  title={{Self-supervised learning from images with a joint-embedding predictive architecture}},
  author={Assran, Mahmoud and Duval, Quentin and Misra, Ishan and Bojanowski, Piotr and Vincent, Pascal and Rabbat, Michael and LeCun, Yann and Ballas, Nicolas},
  booktitle=CVPR,
  year={2023}
}

@inproceedings{dedup,
  author={Aghabagherloo, Alireza and Abadi, Aydin and Sarkar, Sumanta and Dasu, Vishnu Asutosh and Preneel, Bart},
  booktitle={IEEE Security and Privacy Workshops (SPW)}, 
  title={{Impact of Data Duplication on Deep Neural Network-Based Image Classifiers: Robust vs. Standard Models}}, 
  year={2025}
}

@inproceedings{abbas2023semdedup,
  title={{SemDeDup: Data-efficient learning at web-scale through semantic deduplication}},
  author={Abbas, Amro Kamal Mohamed and Tirumala, Kushal and Simig, Daniel and Ganguli, Surya and Morcos, Ari S},
  booktitle={ICLR 2023 Workshop on Mathematical and Empirical Understanding of Foundation Models},
  year={2023}
}

@inproceedings{
dosovitsky2021vit,
title={{An Image is Worth 16x16 Words: Transformers for Image Recognition at Scale}},
author={Alexey Dosovitskiy and Lucas Beyer and Alexander Kolesnikov and Dirk Weissenborn and Xiaohua Zhai and Thomas Unterthiner and Mostafa Dehghani and Matthias Minderer and Georg Heigold and Sylvain Gelly and Jakob Uszkoreit and Neil Houlsby},
booktitle=ICLR,
year={2021},
}

@inproceedings{silberman2012nyu,
  title={Indoor segmentation and support inference from rgbd images},
  author={Silberman, Nathan and Hoiem, Derek and Kohli, Pushmeet and Fergus, Rob},
  booktitle=ECCV,
  year={2012}
}

@inproceedings{geiger2012kitti,
  author = {Andreas Geiger and Philip Lenz and Raquel Urtasun},
  title = {{Are we ready for Autonomous Driving? The KITTI Vision Benchmark Suite}},
  booktitle = CVPR,
  year = {2012}
}

@inproceedings{Cordts2016Cityscapes,
title={{The Cityscapes Dataset for Semantic Urban Scene Understanding}},
author={Cordts, Marius and Omran, Mohamed and Ramos, Sebastian and Rehfeld, Timo and Enzweiler, Markus and Benenson, Rodrigo and Franke, Uwe and Roth, Stefan and Schiele, Bernt},
booktitle=CVPR,
year={2016}
}

@inproceedings{zhou2017ade,
  title={Scene parsing through ade20k dataset},
  author={Zhou, Bolei and Zhao, Hang and Puig, Xavier and Fidler, Sanja and Barriuso, Adela and Torralba, Antonio},
  booktitle=CVPR,
  year={2017}
}

@inproceedings{ranftl2021vision,
  title={Vision transformers for dense prediction},
  author={Ranftl, Ren{\'e} and Bochkovskiy, Alexey and Koltun, Vladlen},
  booktitle=ICCV,
  year={2021}
}

@INPROCEEDINGS{jia2009imagenet,
  author={Deng, Jia and Dong, Wei and Socher, Richard and Li, Li-Jia and Kai Li and Li Fei-Fei},
  booktitle=CVPR, 
  title={{ImageNet: A large-scale hierarchical image database}}, 
  year={2009},
}

@inproceedings{he2020momentum,
  title={Momentum contrast for unsupervised visual representation learning},
  author={He, Kaiming and Fan, Haoqi and Wu, Yuxin and Xie, Saining and Girshick, Ross},
  booktitle=CVPR,
  year={2020}
}

@article{chen2020improved,
  title={Improved baselines with momentum contrastive learning},
  author={Chen, Xinlei and Fan, Haoqi and Girshick, Ross and He, Kaiming},
  journal={arXiv preprint arXiv:2003.04297},
  year={2020}
}

@article{han2025uniquelivessharedworld,
      title={{Unique Lives, Shared World: Learning from Single-Life Videos}}, 
      author={Tengda Han and Sayna Ebrahimi and Dilara Gokay and Li Yang Ku and Maks Ovsjanikov and Iva Babukova and Daniel Zoran and Viorica Patraucean and Joao Carreira and Andrew Zisserman and Dima Damen},
      year={2025},
      journal={arXiv:2512.04085}
}

@inproceedings{chen2021empirical,
  title={An empirical study of training self-supervised vision transformers},
  author={Chen, Xinlei and Xie, Saining and He, Kaiming},
  booktitle=ICCV,
  year={2021}
}

@inproceedings{ridnik2021imagenet21k,
 author = {Ridnik, Tal and Ben-Baruch, Emanuel and Noy, Asaf and Zelnik, Lihi},
 booktitle = {Proceedings of the Neural Information Processing Systems Track on Datasets and Benchmarks},
 title = {{ImageNet-21K Pretraining for the Masses}},
 year = {2021}
}

@article{yang2024depth,
  title={Depth anything v2},
  author={Yang, Lihe and Kang, Bingyi and Huang, Zilong and Zhao, Zhen and Xu, Xiaogang and Feng, Jiashi and Zhao, Hengshuang},
  journal=NeurIPS,
  year={2024}
}

@inproceedings{wang2023cut,
  title={Cut and learn for unsupervised object detection and instance segmentation},
  author={Wang, Xudong and Girdhar, Rohit and Yu, Stella X and Misra, Ishan},
  booktitle=CVPR,
  year={2023}
}

@inproceedings{kerssies2024benchmark,
  title={How to benchmark vision foundation models for semantic segmentation?},
  author={Kerssies, Tommie and De Geus, Daan and Dubbelman, Gijs},
  booktitle=CVPRW,
  year={2024}
}

@inproceedings{
loshchilov2018decoupled,
title={{Decoupled Weight Decay Regularization}},
author={Ilya Loshchilov and Frank Hutter},
booktitle=ICLR,
year={2019},
}

@article{
darcet2025cluster,
title={{Cluster and Predict Latents Patches for Improved Masked Image Modeling}},
author={Timoth{\'e}e Darcet and Federico Baldassarre and Maxime Oquab and Julien Mairal and Piotr Bojanowski},
journal=TMLR,
year={2025},
}

@inproceedings{misra2016shuffle,
  title={Shuffle and learn: unsupervised learning using temporal order verification},
  author={Misra, Ishan and Zitnick, C Lawrence and Hebert, Martial},
  booktitle=ECCV,
  year={2016}
}

@inproceedings{han2019video,
  title={Video representation learning by dense predictive coding},
  author={Han, Tengda and Xie, Weidi and Zisserman, Andrew},
  booktitle={Proceedings of the IEEE/CVF international conference on computer vision workshops},
  year={2019}
}

@article{tong2022videomae,
  title={{VideoMAE}: Masked autoencoders are data-efficient learners for self-supervised video pre-training},
  author={Tong, Zhan and Song, Yibing and Wang, Jue and Wang, Limin},
  journal=NeurIPS,
  year={2022}
}

@inproceedings{wang2023videomaev2,
  title={{VideoMAE v2}: Scaling video masked autoencoders with dual masking},
  author={Wang, Limin and Huang, Bingkun and Zhao, Zhiyu and Tong, Zhan and He, Yinan and Wang, Yi and Wang, Yali and Qiao, Yu},
  booktitle=CVPR,
  year={2023}
}

@article{zhuang2022well,
  title={How well do unsupervised learning algorithms model human real-time and life-long learning?},
  author={Zhuang, Chengxu and Xiang, Ziyu and Bai, Yoon and Jia, Xiaoxuan and Turk-Browne, Nicholas and Norman, Kenneth and DiCarlo, James J and Yamins, Dan},
  journal=NeurIPS,
  year={2022}
}

@inproceedings{purushwalkam2022challenges,
  title={The challenges of continuous self-supervised learning},
  author={Purushwalkam, Senthil and Morgado, Pedro and Gupta, Abhinav},
  booktitle=ECCV,
  year={2022}
}

@article{vitter1985random,
  title={Random sampling with a reservoir},
  author={Vitter, Jeffrey S},
  journal={ACM Transactions on Mathematical Software (TOMS)},
  year={1985}
}

@inproceedings{chen2021exploring,
  title={Exploring simple siamese representation learning},
  author={Chen, Xinlei and He, Kaiming},
  booktitle=CVPR,
  year={2021}
}

@inproceedings{chen2020simple,
  title={A simple framework for contrastive learning of visual representations},
  author={Chen, Ting and Kornblith, Simon and Norouzi, Mohammad and Hinton, Geoffrey},
  booktitle=ICML,
  year={2020}
}

@inproceedings{he2016deep,
  title={Deep residual learning for image recognition},
  author={He, Kaiming and Zhang, Xiangyu and Ren, Shaoqing and Sun, Jian},
  booktitle=CVPR,
  year={2016}
}

@inproceedings{salehi2025mosic,
  title={{MoSiC: Optimal-Transport Motion Trajectory for Dense Self-Supervised Learning}},
  author={Salehi, Mohammadreza and Venkataramanan, Shashanka and Simion, Ioana and Gavves, Efstratios and Snoek, Cees GM and Asano, Yuki M},
  booktitle=ICCV,
  year={2025}
}

@inproceedings{salehi2023time,
  title={Time does tell: Self-supervised time-tuning of dense image representations},
  author={Salehi, Mohammadreza and Gavves, Efstratios and Snoek, Cees GM and Asano, Yuki M},
  booktitle=ICCV,
  year={2023}
}

@inproceedings{siammae2023neurips,
 author = {Gupta, Agrim and Wu, Jiajun and Deng, Jia and Li, Fei-Fei},
 booktitle =NeurIPS,
 title = {{Siamese Masked Autoencoders}},
 year = {2023}
}

@inproceedings{rlt2024neurips,
 author = {Choudhury, Rohan and Zhu, Guanglei and Liu, Sihan and Niinuma, Koichiro and Kitani, Kris M. and Jeni, L\'{a}szl\'{o} A.},
 booktitle =NeurIPS,
 title = {{Don\textquotesingle t Look Twice: Faster Video Transformers with Run-Length Tokenization}},
 year = {2024}
}

@inproceedings{hayes2019memory,
  title={Memory efficient experience replay for streaming learning},
  author={Hayes, Tyler L and Cahill, Nathan D and Kanan, Christopher},
  booktitle={ICRA},
  year={2019}
}

@inproceedings{hayes2020remind,
  title={Remind your neural network to prevent catastrophic forgetting},
  author={Hayes, Tyler L and Kafle, Kushal and Shrestha, Robik and Acharya, Manoj and Kanan, Christopher},
  booktitle=ECCV,
  year={2020},
}

@inproceedings{fini2022self,
  title={Self-supervised models are continual learners},
  author={Fini, Enrico and Da Costa, Victor G Turrisi and Alameda-Pineda, Xavier and Ricci, Elisa and Alahari, Karteek and Mairal, Julien},
  booktitle=CVPR,
  year={2022}
}

@inproceedings{
hu2022how,
title={{How Well Does Self-Supervised Pre-Training Perform with Streaming Data?}},
author={Dapeng Hu and Shipeng Yan and Qizhengqiu Lu and Lanqing HONG and Hailin Hu and Yifan Zhang and Zhenguo Li and Xinchao Wang and Jiashi Feng},
booktitle=ICLR,
year={2022},
}

@inproceedings{marczak2024magmax,
  title={Magmax: Leveraging model merging for seamless continual learning},
  author={Marczak, Daniel and Twardowski, Bart{\l}omiej and Trzci{\'n}ski, Tomasz and Cygert, Sebastian},
  booktitle=ECCV,
  year={2024}   
}

@inproceedings{
roth2024a,
title={{A Practitioner's Guide to Continual Multimodal Pretraining}},
author={Karsten Roth and Vishaal Udandarao and Sebastian Dziadzio and Ameya Prabhu and Mehdi Cherti and Oriol Vinyals and Olivier J Henaff and Samuel Albanie and Matthias Bethge and Zeynep Akata},
booktitle={NeurIPS 2024 Workshop on Scalable Continual Learning for Lifelong Foundation Models},
year={2024},
}

@article{wang2025testtime,
  author  = {Renhao Wang and Yu Sun and Arnuv Tandon and Yossi Gandelsman and Xinlei Chen and Alexei A. Efros and Xiaolong Wang},
  title   = {{Test-Time Training on Video Streams}},
  journal = JMLR,
  year    = {2025}
}

@inproceedings{singh2016krishnacam,
  title={{KrishnaCam}: Using a longitudinal, single-person, egocentric dataset for scene understanding tasks},
  author={Singh, Krishna Kumar and Fatahalian, Kayvon and Efros, Alexei A},
  booktitle={WACV},
  year={2016}
}

@InProceedings{perret2025hdepic,
    author    = {Perrett, Toby and Darkhalil, Ahmad and Sinha, Saptarshi and Emara, Omar and Pollard, Sam and Parida, Kranti Kumar and Liu, Kaiting and Gatti, Prajwal and Bansal, Siddhant and Flanagan, Kevin and Chalk, Jacob and Zhu, Zhifan and Guerrier, Rhodri and Abdelazim, Fahd and Zhu, Bin and Moltisanti, Davide and Wray, Michael and Doughty, Hazel and Damen, Dima},
    title     = {{HD-EPIC: A Highly-Detailed Egocentric Video Dataset}},
    booktitle = CVPR,
    year      = {2025},
}

@article{alam2026global,
  title={A global dataset of continuous urban dashcam driving},
  author={Alam, Md Shadab and Bazilinska, Olena and Bazilinskyy, Pavlo},
  journal={arXiv:2604.01044},
  year={2026}
}

@InProceedings{huang2016droppath,
author={Huang, Gao
and Sun, Yu
and Liu, Zhuang
and Sedra, Daniel
and Weinberger, Kilian Q.},
title={{Deep Networks with Stochastic Depth}},
booktitle=ECCV,
year={2016}
}

@inproceedings{psomas2026attention,
  title={{Attention, please! Revisiting attentive probing through the lens of efficiency}},
  author={Psomas, Bill and Christopoulos, Dionysios and Baltzi, Eirini and Kakogeorgiou, Ioannis and Aravanis, Tilemachos and Komodakis, Nikos and Karantzalos, Konstantinos and Avrithis, Yannis and Tolias, Giorgos},
  booktitle=ICLR,
  year={2026}
}

@inproceedings{fini2025multimodal,
  title={{Multimodal Autoregressive Pre-training of Large Vision Encoders}},
  author={Fini, Enrico and Shukor, Mustafa and Li, Xiujun and Dufter, Philipp and Klein, Michal and Haldimann, David and Aitharaju, Sai and da Costa, Victor G Turrisi and B{\'e}thune, Louis and Gan, Zhe and others},
  booktitle=CVPR,
  year={2025}
}

@inproceedings{mall2025cram,
  title={Cram: Large-scale video continual learning with bootstrapped compression},
  author={Mall, Shivani and Henriques, Joao F},
  booktitle=ICCV,
  year={2025}
}


\appendix
\section{Additional Experiments}

\begingroup\color{changes}\subsection{Intra- and Inter-Batch Similarity on \texttt{WT++12h}}
\label{sec:appendix_wt_similarity}

In \Cref{tab:imagenet_streaming}, we show that high inter-batch similarity
does not explain why the streaming MAE baseline 
lags behind i.i.d.-trained MAE.
A pre-shuffled
ImageNet-1K stream remains close to standard i.i.d. MAE despite its high
inter-batch similarity. We repeat this analysis using exactly the same 
\wtp{} frames and evaluate
both ViT-S/16 and ViT-B/16. We compare standard i.i.d. sampling, a
pre-shuffled stream in which frames are shuffled once before applying
the sliding-window protocol, and the original chronological stream.

\begin{table}[h!]
    \color{changes}
    \centering
    \footnotesize
    \caption{
Intra-/inter-batch similarity and downstream performance for ViT-S/16 and ViT-B/16 pretrained on \wtp{}.
}
    \label{tab:wt_similarity_control}
    \setlength{\tabcolsep}{6pt}
        \begin{tabular}{@{}lcccccc@{}}
        \toprule
        \multirow{2}{*}{Training regime}
        & \multirow{2}{*}{$\mu_{\mathrm{intra}}$}
        & \multirow{2}{*}{$\mu_{\mathrm{inter}}$}
        & \multicolumn{2}{c}{ViT-S/16}
        & \multicolumn{2}{c}{ViT-B/16} \\
        \cmidrule(lr){4-5}
        \cmidrule(lr){6-7}
        & & &
        \makecell[c]{ADE20K \\ (mIoU)~$\uparrow$} &
        \makecell[c]{Cityscapes \\ (mIoU)~$\uparrow$} &
        \makecell[c]{ADE20K \\ (mIoU)~$\uparrow$} &
        \makecell[c]{Cityscapes \\ (mIoU)~$\uparrow$} \\
        \midrule
        i.i.d.
        & 0.171 & 0.738
        & 25.9 & 63.5
        & \textbf{32.9} & 67.8 \\
        Pre-shuffled stream
        & 0.171 & 0.999
        & \textbf{26.0} & \textbf{63.9}
        & 32.3 & \textbf{68.8} \\
        \midrule
        Chronological stream
        & 0.665 & 1.000
        & 23.9 & 61.3
        & 26.2 & 58.2 \\
        \bottomrule
    \end{tabular}
\end{table}

The key comparison is between the two streaming regimes, which have
nearly identical inter-batch similarities of $0.999$ and $1.000$.
Pre-shuffling reduces intra-batch similarity from $0.665$ to $0.171$
and improves ADE20K and Cityscapes by $2.1$ and $2.6$ mIoU with
ViT-S/16 and by $6.1$ and $10.6$ mIoU with ViT-B/16. Moreover,
i.i.d. sampling and pre-shuffled streaming remain comparable despite
their substantially different inter-batch similarities. Together with
the ImageNet-1K control, these results support high intra-batch
similarity as the main source of degradation in a streaming setting.

\subsection{Downstream Evaluation Reliability}
\label{sec:finetuning_reliability}

To assess downstream evaluation reliability for the MAE-based models in the main comparisons reported in \Cref{tab:benchmarking_streaming_methods,tab:tab_scaling_the_encoder_size}, we repeat ImageNet-1K fine-tuning with two seeds and ADE20K and Cityscapes fine-tuning with three seeds.

\begin{table}[h!]
\color{changes}
    \centering
    \footnotesize
    \caption{
    Results across repeated downstream fine-tuning runs for fixed
    ViT-S/16 and ViT-B/16 checkpoints pretrained on \wtp{}.
    We report mean and standard deviation over two seeds for ImageNet-1K
    and three seeds for ADE20K and Cityscapes.
    }
    \label{tab:finetuning_reliability}
        \begin{tabular}{@{}lcccccc@{}}
        \toprule
        \multirow{2}{*}{Method}
        & \multicolumn{3}{c}{ViT-S/16}
        & \multicolumn{3}{c}{ViT-B/16} \\
        \cmidrule(lr){2-4}
        \cmidrule(lr){5-7}
        & \makecell[c]{IN-1K \\ Acc@1~$\uparrow$}
        & \makecell[c]{ADE20K \\ mIoU~$\uparrow$}
        & \makecell[c]{Cityscapes \\ mIoU~$\uparrow$}
        & \makecell[c]{IN-1K \\ Acc@1~$\uparrow$}
        & \makecell[c]{ADE20K \\ mIoU~$\uparrow$}
        & \makecell[c]{Cityscapes \\ mIoU~$\uparrow$} \\
        \midrule
        MAE -- standard i.i.d.
        & \mstd{77.02}{0.02}
        & \mstd{{25.9}}{{0.1}}
        & \mstd{{63.4}}{{0.7}}
        & \mstd{{81.12}}{{0.10}}
        & \mstd{32.8}{0.1}
        & \mstd{68.1}{0.4} \\
        MAE (streaming)
        & \mstd{77.02}{0.09}
        & \mstd{24.1}{0.2}
        & \mstd{60.6}{0.6}
        & \mstd{80.05}{0.04}
        & \mstd{26.6}{0.2}
        & \mstd{58.9}{0.5} \\
        \ours{}
        & \mstd{{77.56}}{{0.05}}
        & \mstd{25.8}{0.3}
        & \mstd{63.3}{0.6}
        & \mstd{{81.12}}{{0.00}}
        & \mstd{{32.9}}{{0.6}}
        & \mstd{{69.5}}{{0.4}} \\
        \bottomrule
    \end{tabular}    
\end{table}

As shown in \Cref{tab:finetuning_reliability}, the improvements over the
streaming MAE baseline are larger than the observed fine-tuning variation for
both backbones. \ours{} also matches or exceeds same-data i.i.d. MAE
across the evaluated tasks. These evaluations preserve the
conclusions of the main comparison.

\endgroup

\subsection{Sensitivity to Batch Size} \label{sec:sensitivity_to_batch_size}
We test how far the temporal extent of each update can be reduced by decreasing the batch size without using \text{DataDrop}. ViT-B/16 pretrained on \wtppp{} with subsampling factor $k=16$ (i.e., \texttt{3.75 FPS}), $B=512$, $256$, and $128$ correspond to roughly 136, 68, and 34 seconds of video per batch. Table~\ref{tab:reducing_bs} shows that performance degrades gradually, but remains competitive even at $B=128$, despite each update containing a very short and highly correlated segment of the stream.
\begin{table}[h!]
    \centering
    \caption{
Effect of reducing batch duration without \text{DataDrop}. With ViT-B/16 pretrained on \wtppp{} at 3.75 FPS, smaller batches correspond to shorter and more temporally correlated video windows, yet performance degrades only gradually.
}
    \setlength{\tabcolsep}{4.5pt}
    \footnotesize
    \label{tab:reducing_bs}
    \setlength{\tabcolsep}{3.2pt}
\begin{tabular}{lccccc}
    \toprule
    \multirow{2}{*}{Method} &
    \multirow{2}{*}{\makecell[c]{Batch size}}  &
    \multirow{2}{*}{\makecell[c]{\texttt{DataDrop}}}  &
    \multirow{2}{*}{\makecell[c]{IN-1K \\ Acc@1~$\uparrow$}} &
    \multirow{2}{*}{\makecell[c]{CS \\ (mIoU)~$\uparrow$}} &
    \multirow{2}{*}{\makecell[c]{ADE \\ (mIoU)~$\uparrow$}} \\
    & & & & & \\
    \midrule
    \multirow{3}{*}{\makecell[l]{\ours{} (ViT-B)}}
        &  \cellcolor{blue!5} $B=512$ &  \cellcolor{blue!5}\no &  \cellcolor{blue!5}81.7 &  \cellcolor{blue!5}73.6 &  \cellcolor{blue!5}36.4 \\
        & $B=256$ & \no & 81.8 & 73.3 & 35.4 \\
        & $B=128$ & \no & 81.4 & 72.2 & 34.2 \\
    \bottomrule
\end{tabular}
\end{table}

\subsection{On Baselines and \text{DataDrop}} \label{subsec:onbaselines_and_datadrop}

We use \text{DataDrop} for the streaming baselines reported in the main paper because it improves their downstream performance. In this setting, we form a larger local stream window with $B=2048$, but backpropagate only through a randomly selected $25\%$ subset, giving an effective batch size of 512. As shown in Table~\ref{tab:on_baselines_and_datadrop}, using the full $B=2048$ window without dropping often hurts performance, while \text{DataDrop} generally matches or improves the $B=512$ setting. We therefore report baseline results with \text{DataDrop} in Figure~\ref{fig:teaser} and Tables~\ref{tab:benchmarking_streaming_methods} and~\ref{tab:tab_scaling_the_encoder_size}. For the gradient-similarity analysis in Figure~\ref{fig:gradient_similarity_figure}, we instead use $B=512$ without \text{DataDrop} for all methods, so that the measured gradients are not affected by example dropping.

\begin{table}[h!]
    \centering
    \footnotesize
    \caption{
    Effect of \text{DataDrop} on streaming baselines. We form a local window of $B=2048$ samples and backpropagate through a randomly selected $25\%$ subset. This generally improves over using the full window and matches or improves the $B=512$ setting.
    }
    \label{tab:on_baselines_and_datadrop}
    \setlength{\tabcolsep}{2.2pt}
\begin{tabular}{lccccc}
    \toprule
    \multirow{2}{*}{Method} &
    \multirow{2}{*}{\makecell[c]{Batch \\ size}}  &
    \multirow{2}{*}{\makecell[c]{\texttt{DataDrop} \\ \texttt{(75\%)}}}  &
    \multirow{2}{*}{\makecell[c]{IN-1K \\ Acc@1~$\uparrow$}} &
    \multirow{2}{*}{\makecell[c]{CS \\ (mIoU)~$\uparrow$}} &
    \multirow{2}{*}{\makecell[c]{ADE \\ (mIoU)~$\uparrow$}} \\
    & & & & & \\
    \midrule
    \multirow{2}{*}{\makecell[l]{MOCOv3~\cite{chen2021empirical} baseline (ViT-S)}}
& $B=2048$ & \no  & 66.3 & 51.7 & 18.6 \\
& $B=2048$ & \yes & 68.7 & 52.0 & 19.4 \\
    \midrule
    \multirow{3}{*}{\makecell[l]{MAE~\cite{he2022masked} baseline w/ Orthogonal-AdamW~\cite{han2025learning} (ViT-S)}}
        & $B=2048$ & \no &  75.2 &  55.3  &  20.6 \\
        & $B=512$ &  \no &  75.5 &  58.6  &  21.5 \\
        & $B=2048$ & \yes & 75.9 &  58.3  & 22.5 \\
    
    \midrule
    \multirow{3}{*}{\makecell[l]{MAE~\cite{he2022masked} baseline (ViT-S)}}
        & $B=2048$ & \no &  76.6 & 57.9 & 23.2 \\
        & $B=512$ &  \no &  76.7 & 60.8 & 23.2 \\
        & $B=2048$ & \yes & 77.1 & 61.3  & 23.9 \\
    
    \bottomrule
\end{tabular}
\end{table}

\subsection{Controlling for the Final Video}

Since streaming runs use a constant learning rate, a possible concern is that gains from longer streams are driven mainly by the last video seen during training. To test this, we append the same final video, \texttt{WT++Budapest}, to shorter streams and compare against the full \wtppp{} stream, where \texttt{Budapest} already appears as the last segment.

\begin{table}[h!]
    \centering
    \footnotesize
    \caption{
    Controlling for the final-video effect. Appending the same final video to shorter streams does not match the full \wtppp{} stream.
    }
    \label{tab:importance_of_last_video}
    \setlength{\tabcolsep}{3.2pt}
\begin{tabular}{l l ccc}
    \toprule
    \multirow{2}{*}{Method} &
    \multirow{2}{*}{\texttt{WT} duration}  &
    \multirow{2}{*}{\makecell[c]{IN-1K \\ Acc@1~$\uparrow$}} &
    \multirow{2}{*}{\makecell[c]{CS \\ (mIoU)~$\uparrow$}} &
    \multirow{2}{*}{\makecell[c]{ADE \\ (mIoU)~$\uparrow$}} \\
    & & & & \\
    \midrule
    \multirow{3}{*}{\makecell[l]{\ours{} (ViT-B)}}
 
        & \wtp{} \texttt{+Budapest}  & \text{81.1} & \text{69.4} & \text{32.5} \\
        & \wtpp{} \texttt{+Budapest} & \text{81.5} & \text{72.2} & \text{35.2} \\
        & \wtppp{}                  & \text{81.9} & \text{73.8} & \text{35.6} \\
    \bottomrule
\end{tabular}
\end{table}

As shown in Table~\ref{tab:importance_of_last_video}, appending \texttt{Budapest} does not match training on the full \wtppp{} stream. This suggests that the gains from longer pretraining are not explained solely by the most recent video, but also depend on the preceding stream.

\subsection{Masking Strategy}\label{sec:appendix_masking_strategy}
We also evaluate whether alternative masking strategies improve \ours{}. Table~\ref{tab:different_masking_strategy} compares random uniform masking with block masking~\cite{yang2025pixio} and masking inspired by I-JEPA~\cite{assran2023ijepa}/Bootleg~\cite{lowe2026bootleg}. We find that random uniform masking remains the strongest overall choice: block masking performs comparably on ADE20K, but underperforms on ImageNet-1K and Cityscapes, while the I-JEPA/Bootleg-style strategy is consistently worse. We therefore use standard random uniform masking in our experiments.

\begin{table}[h!]
    \centering
    \footnotesize
    \caption{
    Effect of masking strategy for \ours{} with ViT-B/16. Standard random uniform masking remains the strongest overall choice.
    }
    \label{tab:different_masking_strategy}
    \setlength{\tabcolsep}{3.2pt}
\begin{tabular}{l l ccc}
    \toprule
    \multirow{2}{*}{Method} &
    \multirow{2}{*}{Masking Strategy} &
    \multirow{2}{*}{\makecell[c]{IN-1K \\ Acc@1~$\uparrow$}} &
    \multirow{2}{*}{\makecell[c]{CS \\ (mIoU)~$\uparrow$}} &
    \multirow{2}{*}{\makecell[c]{ADE \\ (mIoU)~$\uparrow$}} \\
    & & & & \\
    \midrule
    \multirow{3}{*}{\makecell[l]{\ours{} (ViT-B)}}
        & Block Masking~\cite{yang2025pixio} (75\%) 
        & \text{80.9} & \text{68.1} & \text{32.8} \\
        & I-JEPA~\cite{assran2023ijepa} / Bootleg~\cite{lowe2026bootleg}
        & \text{80.4} & \text{66.9} & \text{31.7} \\
        & \cellcolor{blue!5} Random Uniform~\cite{he2022masked} (75\%)
        & \cellcolor{blue!5} \text{81.1}
        & \cellcolor{blue!5} \text{69.0}
        & \cellcolor{blue!5} \text{32.7} \\
    \bottomrule
\end{tabular}
\end{table}

\subsection{AdamW Momentum}\label{sec:appendix_adamw_momentum}

Following \citet{carreira2024learning}, we evaluate AdamW~\cite{loshchilov2018decoupled} without first-moment momentum by setting $\beta_1=0$. As shown in Table~\ref{tab:zeroing_adamw_momentum}, this setting underperforms our default AdamW configuration with $\beta_1=0.9$ for \ours{} with ViT-B. We therefore use $\beta_1=0.9$ throughout the paper.

\begin{table}[h!]
    \centering
    \footnotesize
    \caption{
    Effect of AdamW first-moment momentum for \ours{} with ViT-B/16. Our standard setting $\beta_1=0.9$ performs better than deactivating momentum.
    }
    \label{tab:zeroing_adamw_momentum}
    \begin{tabular}{l l ccc}
    \toprule
    \multirow{2}{*}{Method} &
    \multirow{2}{*}{AdamW $\beta_1$} &
    \multirow{2}{*}{\makecell[c]{IN-1K \\ Acc@1~$\uparrow$}} &
    \multirow{2}{*}{\makecell[c]{CS \\ (mIoU)~$\uparrow$}} &
    \multirow{2}{*}{\makecell[c]{ADE \\ (mIoU)~$\uparrow$}} \\
    & & & & \\
    \midrule
    \multirow{2}{*}{\makecell[l]{\ours{} (ViT-B)}}
        & $0$
        & \text{81.0} & \text{67.1} & \text{31.8} \\
        &  \cellcolor{blue!5}$0.9$
        & \cellcolor{blue!5} \text{81.1}
        & \cellcolor{blue!5} \text{69.0}
        & \cellcolor{blue!5} \text{32.7} \\
    \bottomrule
\end{tabular}
\end{table}

\subsection{Additional Ablations}
\begin{table}[h!]
    \centering
    \footnotesize
    \caption{
Additional ablation of \ours{} components with ViT-S/16 pretrained on \wtp{}. The extra rows remove either \text{DataDrop} or regularization from the full method to isolate their roles.
}
    \label{tab:additional_ablations_appendix}
    \begin{tabular}{lccccccc}
        \toprule
        & \multicolumn{4}{c}{Components} & \multicolumn{3}{c}{Downstream performance} \\
        \cmidrule(lr){2-5} \cmidrule(lr){6-8}
        &
        \multirow{2}{*}{\texttt{DataDrop}} &
        \multirow{2}{*}{\makecell[c]{Color Jit. \& \\ Drop Path}} &
        \multirow{2}{*}{\makecell[c]{Two-stage \\ Cropping}} &
        \multirow{2}{*}{\makecell[c]{Motion-biased \\ Crop Selection}} &
        \multirow{2}{*}{Cityscapes $\uparrow$} & \multirow{2}{*}{ADE20K $\uparrow$} & \multirow{2}{*}{KITTI $\downarrow$} \\
        &&&&&&& \\
        
        \midrule
            & -- & -- & -- &                          --           & 60.8 & 23.2  & \mstd{4.307}{0.046} \\
            & \checkmark & -- & -- &                  --          & 61.3 & 23.9   & \mstd{4.211}{0.137} \\
            & \checkmark & \checkmark & -- &          --          & 62.6 & 25.1  & \mstd{4.025}{0.039} \\
            & \checkmark & \checkmark & \checkmark                & -- & 62.8 & 26.1 & \mstd{4.074}{0.048} \\
            & \checkmark & -- & \checkmark &  \checkmark          & 62.4 & 24.7  & \mstd{4.022}{0.123} \\
            & -- & \checkmark & \checkmark &  \checkmark          & 63.1 & 25.4  & \mstd{\mathbf{3.958}}{\mathbf{0.087}} \\
            & \checkmark & \checkmark & \checkmark &  \checkmark  & \textbf{63.8} & \textbf{26.1}  & \mstd{3.976}{0.012} \\
        \bottomrule
   \end{tabular}

\end{table}
Table~\ref{tab:additional_ablations_appendix} extends the main ablation with two additional configurations that remove either \text{DataDrop} or the combination of color jitter and increased drop path from the full method, while keeping crop selection fixed. Removing color jitter and increased drop path lowers performance, confirming their importance under low within-batch diversity. Removing \text{DataDrop} remains competitive on semantic segmentation, suggesting that \text{DataDrop} is a useful complementary component but not the main driver of \ours{}. This is consistent with our scaling experiments, where \ours{} remains effective even without \text{DataDrop} when using larger models and longer streams.

\begin{table}[h!]
\color{changes}
    \centering
    \footnotesize
    \caption{
    Ablation of motion-biased crop selection with ViT-B/16 pretrained on
    KrishnaCAM. Both variants use the same frames, pretraining budget, and
    remaining \ours{} configuration.
    }
    \label{tab:krishnacam_motion_crops}
    \setlength{\tabcolsep}{2.5pt}
    \renewcommand{\arraystretch}{1.05}
        \begin{tabular}{@{}lcccc@{}}
        \toprule
        Method
        & \makecell[c]{ADE \\ (mIoU)~$\uparrow$}
        & \makecell[c]{CS \\ (mIoU)~$\uparrow$}
        & \makecell[c]{NYUv2 \\ (RMSE)~$\downarrow$}
        & \makecell[c]{KITTI \\ (RMSE)~$\downarrow$} \\
        \midrule
        \ours{} w/o motion-biased selection
        & 33.9
        & 71.5
        & \mstd{\mathbf{0.621}}{\mathbf{0.004}}
        & \mstd{3.646}{0.053} \\
        \ours{} (ours)
        & \textbf{34.5}
        & \textbf{72.2}
        & \mstd{\mathbf{0.621}}{\mathbf{0.001}}
        & \mstd{\mathbf{3.583}}{\mathbf{0.012}} \\
        \bottomrule
    \end{tabular}
\end{table}

\subsection{Longer i.i.d.~ImageNet Reference}
In the main benchmarking table (\Cref{tab:benchmarking_streaming_methods}) the ImageNet-1K i.i.d.~MAE reference is iteration-matched to the \wtp{} streaming runs. Here, we provide an additional reference trained for the same number of iterations as longer \wtpppp{} streaming runs. This corresponds to approximately $650\mathrm{k}$ iterations, matching 95 hours of video at 3.75 FPS with stride $s=2$. This comparison contextualizes the scaled \ours{} results against a longer i.i.d.~ImageNet pretraining budget.
\begin{table}[h!]
    \centering
    \footnotesize
\caption{
Longer i.i.d.~ImageNet reference. Unlike the main benchmarking tables, where ImageNet-1K i.i.d.~MAE is iteration-matched to \wtp{}, this table matches the training length of the \wtpppp{} streaming runs, approximately $650\mathrm{k}$ iterations. \ours{} trained on \wtpppp{} remains competitive with this longer i.i.d.~ImageNet reference, particularly for ViT-S and Cityscapes transfer.
}
    \label{tab:increasing_in1k_iterations}
    \begin{tabular}{l l ccc}
    \toprule
    \multirow{2}{*}{Method} &
    \multirow{2}{*}{Pretraining data} &
    \multirow{2}{*}{\makecell[c]{IN-1K \\ Acc@1~$\uparrow$}} &
    \multirow{2}{*}{\makecell[c]{CS \\ (mIoU)~$\uparrow$}} &
    \multirow{2}{*}{\makecell[c]{ADE \\ (mIoU)~$\uparrow$}} \\
    & & & & \\
    \midrule
    \rowcolor{gray!8} Standard i.i.d. MAE (ViT-S) & ImageNet-1K & 78.5 & 65.9 & 28.5 \\
    \rowcolor{blue!8}\ours{} (ViT-S) & \wtpppp{} & 78.4 & 68.0 & 29.5 \\
    \midrule
    \rowcolor{gray!8} Standard i.i.d. MAE (ViT-B) & ImageNet-1K & 82.5 & 75.2 & 39.8 \\
    \rowcolor{blue!8}\ours{} (ViT-B) & \wtpppp{} & 82.0 & 74.0 & 36.5 \\
    \rowcolor{blue!8}\ours{} (ViT-B, avg. \{60,100\}\%; see~\Cref{tab:checkpoint_averaging}) & \wtpppp{} & 81.9 & 75.3 & 37.3 \\
    \bottomrule
\end{tabular}
\end{table}

\subsection{Cosine Similarity of Gradients from Consecutive Batches}

\begin{figure}[h!]
    \centering
    \includegraphics[width=1.0\linewidth]{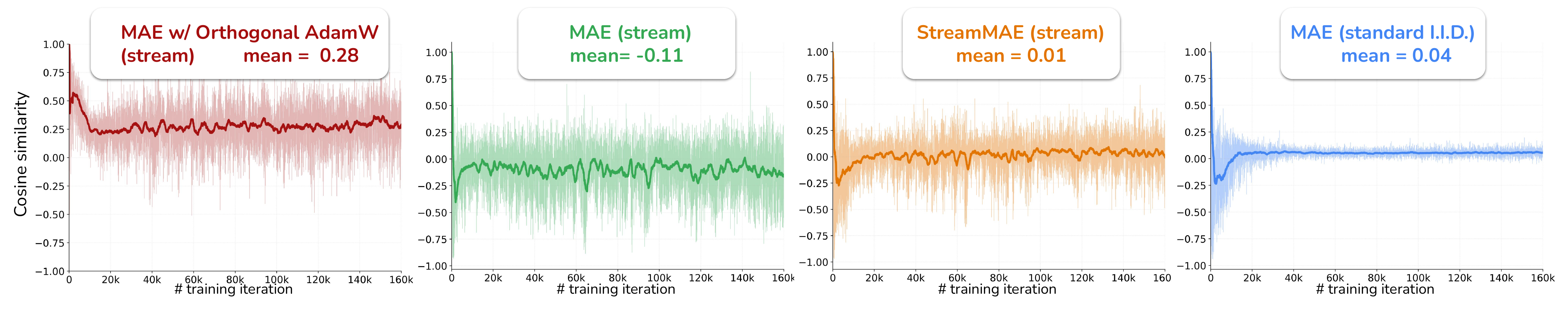}
    \caption{
    Running average of cosine similarity between gradients from consecutive batches, measured on last-block MLP parameters every 10 steps.
    }
    \label{fig:gradient_similarity_figure_appendix}
\end{figure}

In \Cref{fig:gradient_similarity_figure_appendix}, we provide a more detailed visualization of the consecutive-gradient similarity analysis from the main paper. For clarity, we plot each MAE-based method separately and show a running average over 100 logged values. Gradient cosine similarity is logged every 10 training steps using the MLP parameters in the last transformer block. While \ours{} has a mean similarity close to the standard i.i.d.~MAE reference, the i.i.d.~run exhibits lower variance.

\begingroup
\color{changes}
\Cref{fig:mean_sim_vs_performance_appendix} extends the analysis in \Cref{fig:gradient_similarity_figure} with three additional runs: MAE trained on a pre-shuffled \wtp{} stream (\Cref{tab:wt_similarity_control}), \ours{} with Orthogonal-AdamW, and streaming MAE using only motion-biased crop selection. All models use a ViT-S/16 backbone and are pretrained on \wtp{}. The streaming variants use \(B=512\) without \text{DataDrop}. Across the evaluated configurations, a smaller distance from standard i.i.d. MAE in mean consecutive-batch gradient cosine similarity over training is generally associated with stronger downstream performance on Cityscapes and ADE20K. The only clear exception is streaming MAE with motion-biased crop selection on Cityscapes, which is closer to i.i.d. MAE in gradient similarity but performs similarly to the streaming MAE baseline. Notably, MAE trained on the pre-shuffled \wtp{} stream is the closest streaming variant to i.i.d. MAE in gradient similarity and achieves comparable downstream performance. This further supports the findings in \Cref{tab:imagenet_streaming,tab:wt_similarity_control} that high inter-batch similarity does not explain the degradation observed under streaming pretraining.

\begin{figure}
    \color{changes}
    \centering
    \includegraphics[width=\linewidth]{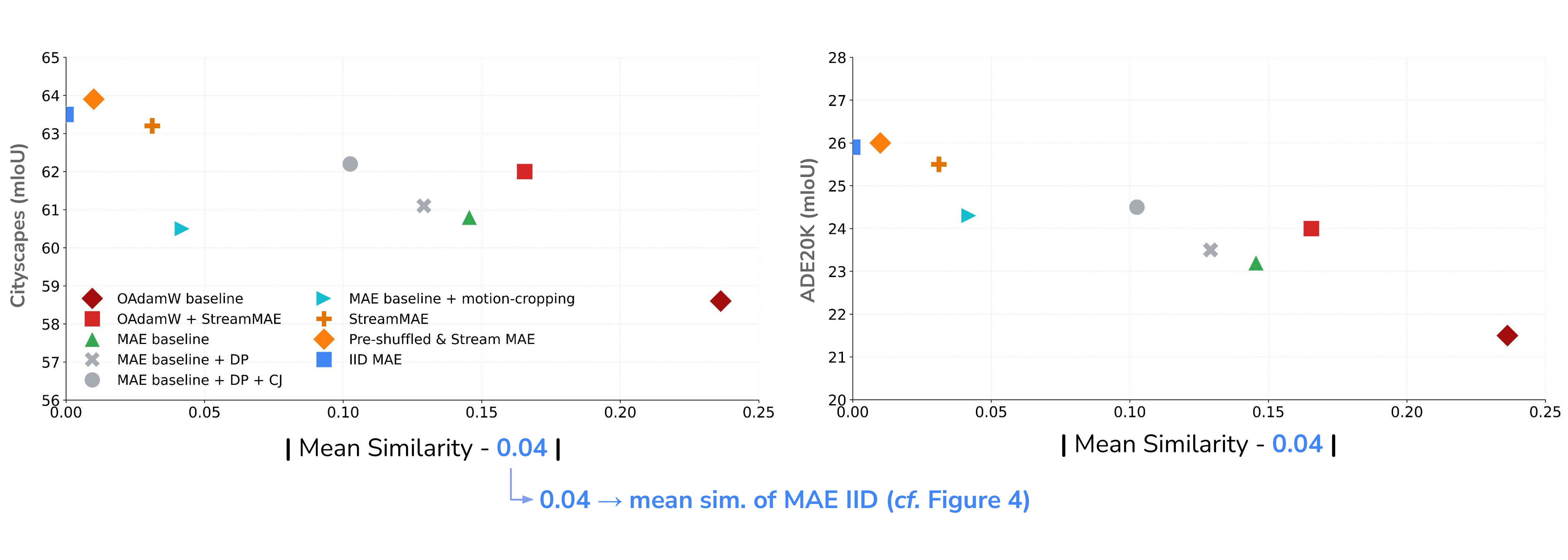}
    \caption{
Downstream performance versus the distance from standard i.i.d. MAE in mean gradient cosine similarity between consecutive batches throughout training. Extending \Cref{fig:gradient_similarity_figure}, we include MAE trained on pre-shuffled \& streaming \wtp{} (\Cref{tab:wt_similarity_control}), \ours{} with Orthogonal-AdamW, and streaming MAE using only motion-biased crop selection. All models use a ViT-S/16 backbone and are pretrained on \wtp{}. All streaming variants use $B=512$ without \text{DataDrop}.
}
    \label{fig:mean_sim_vs_performance_appendix}
\end{figure}
\endgroup

\section{Limitations and Open Challenges} \label{sec:limitiations_and_future_work}

We acknowledge several limitations of our study. First, except for motion-biased crop selection, \ours{} does not explicitly model temporal dynamics. Extending \ours{} to architectures or objectives that directly exploit temporal information remains an important direction for future work. Second, \ours{} relies on fixed frame subsampling to reduce the effective frame rate, motivated by the observation in Table~\ref{tab:increasing_fps} that denser temporal sampling can degrade performance. However, fixed subsampling is likely suboptimal, since the rate of visual change can vary substantially throughout a video, as illustrated in Figure~\ref{fig:analyzing_of_frames_with_dv2_features}. Adaptive temporal sampling may therefore better match the local dynamics of the stream. Finally, we only consider checkpoint averaging at evaluation time, while incorporating long-timescale consolidation directly into streaming pretraining remains unexplored.

\section{\texttt{WT++} dataset} \label{sec:wtppdataset}
\begin{table}[h!]
    \centering
    \footnotesize
    \caption{
    Construction of the ordered \texttt{WT++} streams used in scaling experiments. Each stream is a prefix of the next one.
    }
    \label{tab:wtpp_stream_construction}
    \setlength{\tabcolsep}{4.5pt}
    \begin{tabular}{lccc}
        \toprule
        Stream & \# Videos & Duration  \\
        \midrule
        \wtp{}    & 1  & 12.0h & \texttt{London} stream \\
        \wtpp{}   & 10 & 24.8h & \wtp{} + 9 videos from original WalkingTours~\cite{venkataramanan2024dora} \\
        \wtppp{}  & 26 & 50.0h & \wtpp{} + 16 appended videos \\
        \wtpppp{} & 58 & 94.5h & Full \texttt{WT++} collection \\
        \bottomrule
    \end{tabular}
\end{table}

We provide additional details on the construction of \texttt{WT++}. The dataset contains 58 urban walking-tour videos with a total duration of 94.5 hours. Videos are originally recorded at 60 FPS, similar to the released WalkingTours dataset. In our experiments, we temporally subsample each stream before training; unless stated otherwise, we use subsampling factor $k=16$, corresponding to an effective frame rate of \texttt{3.75 FPS}. For scaling experiments, we construct ordered streams of increasing duration by sequentially appending videos, yielding \wtp{}, \wtpp{}, \wtppp{}, and \wtpppp{}. Each stream extends the preceding one with new videos while retaining all earlier videos in the same order. Thus, scaling pretraining duration introduces new visual content while preserving the earlier part of the stream.

\begin{table}[h!]
    \centering
    \scriptsize
    \caption{
    Construction of the ordered \texttt{WT++} streams. Each row lists the new videos introduced at that stage, in training order (all videos from earlier stages are retained).
    }
    \label{tab:wtpp_video_level}
    \setlength{\tabcolsep}{4.0pt}
    \begin{tabular}{l c p{0.74\linewidth}}
        \toprule
        Stream stage & \# New videos & Cities / videos in training order \\
        \midrule
        \wtp{} & 1 &
        London \\
        \midrule
        \wtpp{} & 9 &
        Venice, Amsterdam, Singapore, Istanbul, Bangkok, Stockholm, Kuala Lumpur, Zurich, Chiang Mai \\
        \midrule
        \wtppp{} & 16 &
        Frankfurt, Phnom Penh, Sevilla, Hoi An, Prague, Chemnitz, Marmaris, Dubrovnik, Mostar, Gibraltar, Barcelona, Valencia, Rhodes, Helsinki, Copenhagen, Budapest \\
        \midrule
        \wtpppp{} & 32 &
        Vilnius, Danang, Klaipeda, Liege, Marbella, Monschau, Naples, Nessebar, Port de Soller, Sarajevo, Arcadia, Timisoara, Warsaw, Malacca, Florence, Cardiff, Georgetown, Luxembourg, Alicante, Ipoh, Belfast, Kaliningrad, Maastricht, Valletta, Vienna, Saigon, Cadiz, Kyiv, Oxford, Sihanoukville, Chongqing, Tokyo \\
        \bottomrule
    \end{tabular}
\end{table}

\subsection{Dataset release} \label{subsec:dataset_release}
All videos used in \texttt{WT++} are publicly accessible online. We will release the dataset package, including metadata, source links, and preprocessing scripts, under CC BY 4.0. For videos already licensed under CC BY 4.0 (56 out of 58 videos), we will redistribute processed copies under their original CC BY 4.0 terms with attribution to the original creators. For the two videos currently under the Standard YouTube License, we are seeking written permission for direct file redistribution from the creators. If permission is not obtained, we will release only metadata and source links for these videos, or replace them with compatible alternatives.

\section{\textit{MemoryStoryboard}~\cite{yang2026memorystoryboard} Results} \label{sec:memory_storyboard_results}

For the \textit{MemoryStoryboard} baseline, we use the official codebase and replace the default ResNet-50~\cite{he2016deep} encoder with ViT-S/16 to match the architecture used in our main comparisons. The encoder is initialized from scratch, and we use the final \texttt{[CLS]} token as the image representation. We train on the \texttt{WT++London} stream with $224 \times 224$ crops, a short FIFO memory of $2048$ samples, and a long reservoir memory of $16384$ samples. Each batch contains 64 current-stream samples and 448 replay samples, giving an effective batch size of 512. We additionally swept the stream stride and reservoir size and evaluated input resolutions of $112 \times 112$, as used in the original setup, and $224 \times 224$. The reported configuration uses stride $s=2$, which performed better than $s=1$.

As \textit{MemoryStoryboard} uses SimSiam~\cite{chen2021exploring}  and SimCLR~\cite{chen2020simple} objectives, we evaluate both, and tune the learning rate over $\{8 \times 10^{-5}, 10^{-4}, 3 \times 10^{-4}\}$. The best result is obtained with SimCLR and learning rate $10^{-4}$, which is the configuration reported in \Cref{fig:teaser} and \Cref{tab:benchmarking_streaming_methods}. We use AdamW with weight decay $0.05$ and enable the default label-merging procedure.
We also evaluated the original ResNet-50 recipe on \wtp{} using only the most recent $2048$ frames, without long-term replay. Under this constrained memory setting, the learned representations showed signs of collapse, so we do not report downstream results for this configuration.

Since \textit{MemoryStoryboard} was originally developed with ResNet-50 backbones and SimSiam/SimCLR objectives, adapting it to ViT-S/16 may require additional architecture-specific tuning. We therefore view this result as a controlled reproduction under our ViT-S/16 streaming evaluation protocol rather than as an exhaustive optimization of the method. In particular, the original \textit{MemoryStoryboard} setup uses larger strides between consecutive batches, which may be an additional direction for optimizing its performance in our setting.

\section{Additional Training Details}
\label{sec:appendix_training_details}

\subsection{Inter- and Intra-Batch Similarity Computation}
\label{sec:appendix_sims}

We quantify inter- and intra-batch similarity using frozen DINOv2-L~\cite{oquab2023dinov2} features. All similarities are computed on raw images or frames, without training augmentations. Thus, $\mathbf{X}^{(t)}$ denotes the batch/window of underlying samples from the stream, not the augmented views used during SSL training. For each image or frame $\mathbf{x}^{(t)}_i \in \mathbf{X}^{(t)}=\{\mathbf{x}^{(t)}_1,\ldots,\mathbf{x}^{(t)}_B\}$, we extract the \texttt{[CLS]} token representation and $\ell_2$-normalize it:
\[
\mathbf{z}^{(t)}_i =
\frac{f_{\mathrm{DINOv2}}(\mathbf{x}^{(t)}_i)}
{\|f_{\mathrm{DINOv2}}(\mathbf{x}^{(t)}_i)\|_2}.
\]
We define \emph{intra-batch similarity} as the mean off-diagonal cosine similarity within a batch:
\[
\mu_{\mathrm{intra}}(\mathbf{X}^{(t)})
=
\frac{1}{B(B-1)}
\sum_{i\neq j}
(\mathbf{z}^{(t)}_i)^{\top} \mathbf{z}^{(t)}_j
=
\frac{2}{B(B-1)}
\sum_{i<j}
(\mathbf{z}^{(t)}_i)^{\top} \mathbf{z}^{(t)}_j .
\]
This measures how similar examples are within a single optimization batch.
To measure similarity between consecutive batches, we use a symmetric nearest-neighbor score, since duplicate or near-duplicate samples are known to degrade representation learning~\cite{dedup,abbas2023semdedup}. This metric directly estimates whether each example has a highly similar counterpart within the immediately following batch.
For two consecutive batches $\mathbf{X}^{(t)}$ and $\mathbf{X}^{(t+1)}$, we first compute the directed similarities:
\[
\mu_{t\rightarrow t+1}
=
\frac{1}{B}
\sum_{i=1}^{B}
\max_{1\leq j\leq B}
(\mathbf{z}^{(t)}_i)^{\top}\mathbf{z}^{(t+1)}_j,
\]
and
\[
\mu_{t+1\rightarrow t}
=
\frac{1}{B}
\sum_{j=1}^{B}
\max_{1\leq i\leq B}
(\mathbf{z}^{(t+1)}_j)^{\top}\mathbf{z}^{(t)}_i.
\]
The \emph{inter-batch similarity} is the average of the two directions:
\[
\mu_{\mathrm{inter}}(\mathbf{X}^{(t)},\mathbf{X}^{(t+1)})
=
\frac{1}{2}
\left(
\mu_{t\rightarrow t+1}
+
\mu_{t+1\rightarrow t}
\right).
\]
This captures whether examples in one batch have close matches in the next batch.

For the reported values, we average $\mu_{\mathrm{intra}}$ over sampled batches and $\mu_{\mathrm{inter}}$ over consecutive batch pairs. For pre-shuffled and streaming ImageNet-1K, we use batches of size $B=512$ with stride $s=8$, matching the corresponding experiments, and report the mean over three stream starting points, each containing 100 consecutive batches. For \texttt{WT++London}, we use $B=512$ and stride $s=1$, matching our streaming setting, and again average over three random starting points with 100 consecutive batches each. For standard i.i.d.~ImageNet-1K, each batch is sampled independently from the full training set, and here we report the mean over 50 batches, again using batch size $B=512$.

\subsection{Two-Stage Cropping and Motion-Biased Crop Selection}

Videos are originally resized to $1280 \times 720$ resolution. For two-stage cropping, the first-stage crop size is $592 \times 336$, approximately preserving the 16:9 aspect ratio while remaining aligned with the ViT patch grid. The final MAE random resized crop is then applied within this first-stage region. For motion-biased crop selection, we cache frame-difference scores at the ViT patch level. Since the ViT patch size is $16$, each $1280 \times 720$ frame yields a compact $80 \times 45$ patch-level difference map. This cache is inexpensive to store and avoids recomputing frame differences or reloading frames when selecting motion-biased crops. First-stage crop candidates are constrained to be aligned with the ViT patch grid, so their motion scores can be computed directly from the cached patch-level differences. When \text{DataDrop} is used, patch-level differences are still computed for the local stream window before example dropping. Thus, crop selection can use the same cached difference maps, while only the retained examples contribute to the training loss.

\subsection{Training i.i.d.~Baselines}

The i.i.d.~MAE baselines on ImageNet-1K and \texttt{WT++London}, reported in \Cref{tab:imagenet_streaming,tab:benchmarking_streaming_methods,tab:tab_scaling_the_encoder_size}, are trained with an iteration-matched protocol. That is, each i.i.d.~baseline is trained for the same number of optimization steps as the corresponding streaming run on \texttt{WT++London}. These baselines are therefore not intended to reproduce fully optimized MAE pretraining schedules. They provide controlled references at the same training budget as our streaming experiments.

By i.i.d.~training, we mean that each optimization step samples a batch uniformly at random from the full dataset, without preserving temporal order or using sliding-window batches. We train these i.i.d.~baselines with the standard MAE cosine learning-rate schedule, whereas streaming runs use a constant learning rate with linear warm-up of 5\% of total number of iterations. Unless stated otherwise, the batch size is $512$. For IN-1K ViT-B/16 i.i.d.~experiments, we use base learning rate $8\times10^{-5}$ with linear batch-size scaling. For ViT-S/16, we use base learning rate $3.2\times10^{-4}$, which performed better. 

The fixed-order pre-shuffled ImageNet-1K experiment in \Cref{tab:imagenet_streaming} is also iteration-matched. We pre-shuffle ImageNet-1K once, consume the resulting sequence as a fixed stream with batch size $B=512$ and stride $s=8$, and train for approximately the same number of steps as the corresponding i.i.d.~baseline.

\subsection{SSL Baselines}
\label{sec:appendix_baseline_tuning}
We reproduce three representative SSL baselines in our streaming video setup: DINO~\cite{caron2021emerging}, MAE~\cite{he2022masked}, and MoCo v3~\cite{chen2021empirical}. All methods are implemented in solo-learn~\cite{sololearn2022turrisi} and trained from scratch on the \wtp{} stream using the same single-traversal sliding window protocol. Unless stated otherwise, batches are formed in temporal order with stride $s=1$, and baselines use a local stream window of size $B=2048$ with \text{DataDrop}, yielding an effective batch size of 512. Training uses mixed precision, synchronized batch normalization when required by the method, and a constant learning rate with linear warm-up.

\noindent\textbf{DINO~\cite{caron2021emerging}.}
We train DINO with a ViT-S/16 backbone, AdamW, learning rate $10^{-4}$, weight decay $0.04$, and teacher momentum increased from $0.996$ to $0.9996$ with a cosine schedule. We use $8192$ prototypes, projection hidden dimension $2048$, projection output dimension $256$, student temperature $0.1$, and teacher temperature $0.04$. The augmentation pipeline follows the multi-crop recipe with two global $224 \times 224$ crops and four local $96 \times 96$ crops. Increasing the number of prototypes or local crops did not lead to consistent improvements.

\noindent\textbf{MAE~\cite{he2022masked}.}
We train MAE with a ViT-S/16 backbone, AdamW, learning rate $8\times 10^{-5}$, weight decay $0.05$, and AdamW betas $(0.9, 0.95)$. We use the standard masking ratio of $0.75$, normalized pixel loss, and a decoder with embedding dimension $512$, depth $8$, and $16$ attention heads. The MAE baseline uses a single $224 \times 224$ reconstruction crop with random resized cropping and horizontal flipping, without color jitter or blur. We also evaluate MAE with Orthogonal-AdamW~\cite{han2025learning} using the same configuration.

\noindent\textbf{MoCo v3~\cite{chen2021empirical}.}
We train MoCo v3 with a ViT-S/16 backbone, AdamW, learning rate $10^{-4}$, weight decay $0.1$, and teacher momentum increased from $0.996$ to $0.9996$ with a cosine schedule. The projection and prediction hidden dimensions are $4096$, the projection output dimension is $256$, and the contrastive temperature is $0.2$. We use the standard asymmetric two-crop MoCo v3 augmentation pipeline with two $224 \times 224$ crops, color jitter, grayscale augmentation, Gaussian blur, and solarization. Increasing the effective batch size  (i.e., removing \text{DataDrop}) did not improve MoCo v3 in our streaming setup (please see \Cref{tab:on_baselines_and_datadrop}). For \ours{}, we follow the standard MoCo v3 color-jitter setting: brightness $0.4$, contrast $0.4$, saturation $0.2$, and hue $0.1$, and apply it with probability $0.5$.

\begingroup\color{changes}\noindent\textbf{Streaming-specific adaptations.}
We did not assume a priori that MAE would be the strongest objective. We swept the baselines over learning rate, batch size, teacher momentum, crop configuration, and prototype count where applicable. We also tested several adaptations designed specifically for high temporal redundancy.

For \textbf{MoCo v3}, we tested multiple temporal positives, temporally soft targets, slow and fast EMA teachers with different momentum values, and periodic checkpoint merging and reinitialization. None of these variants resolved the optimization instability or prevented downstream transfer from degrading as pretraining progressed.

For \textbf{DINO}, we tested slow and fast EMA teachers, slow- and fast-evolving prototypes, updates restricted to hard examples, and Sinkhorn-Knopp assignment after observing low prototype utilization. None of these variants consistently improved downstream transfer.

We also tested Orthogonal-AdamW with MAE, MoCo v3, and DINO without consistent gains. DataDrop is applied to all streaming baselines in the main comparison because it matches or improves their corresponding no-drop configurations (see \Cref{tab:on_baselines_and_datadrop}). 

\subsection{Additional Pretraining Datasets} \label{sec:appendix_additional_datasets_construction}

We construct three additional ordered pretraining streams from HD-EPIC~\cite{perret2025hdepic}, CROWD~\cite{alam2026global}, and KrishnaCAM~\cite{singh2016krishnacam} (see Tab.~\ref{tab:domain_generalization}). For each dataset, we preserve the temporal order within individual videos and concatenate the selected videos into a single fixed stream.

\textbf{HD-EPIC}~\cite{perret2025hdepic} contains egocentric video recorded entirely in kitchens. We use recordings from four kitchens, P01, P02, P06, and P08, preserving the provided chronological order. The constituent videos average approximately 15 minutes and form an 18-hour stream. We temporally sample the stream at 2.5~FPS, yielding approximately 160k pretraining iterations.

\textbf{CROWD}~\cite{alam2026global} contains front-facing urban dashcam video. We use 16 videos recorded in Edinburgh, Salt Lake City, Jinan, Da Nang, Quito, and Gold Coast. The videos average approximately 53.5 minutes and form a 14.5-hour stream. We temporally sample the stream at 4~FPS, yielding approximately 203k pretraining iterations.

\textbf{KrishnaCAM}~\cite{singh2016krishnacam} is a 70-hour longitudinal egocentric dataset covering indoor and outdoor daily life. Its constituent videos average approximately 9.5 minutes and are concatenated chronologically. We temporally sample the resulting stream at 1~FPS, yielding approximately 250k pretraining iterations. For KrishnaCAM, \ours{} uses $K=2$ candidate crops to reduce repeated selection of the same regions during stationary activities.

\endgroup
\section{Additional Evaluation Details} \label{sec:appendix_evaluation_details}
\subsection{Image Classification}

For ImageNet-1K~\cite{jia2009imagenet} classification, we follow the MAE~\cite{he2022masked} fine-tuning protocol. We fine-tune ViT-S/16 and ViT-B/16 end-to-end at resolution $224 \times 224$, initializing the backbone from the pretrained checkpoint and training a new linear classifier. For MAE-based models, we use global average pooling over patch tokens followed by normalization and a linear classifier; for DINO and MoCo v3, we use the final \texttt{[CLS]} token representation. We fine-tune for 100 epochs with AdamW, cosine learning-rate decay, 5 warmup epochs, weight decay $0.05$, layer-wise learning-rate decay $0.65$, drop-path rate $0.1$, mixup $0.8$, cutmix $1.0$, and random erasing probability $0.25$. We use the learning-rate scaling rule $\mathrm{lr}=\mathrm{blr}\cdot B_{\mathrm{eff}}/256$, with $\mathrm{blr}=5\times 10^{-4}$ and global batch size $B_{\mathrm{eff}}=1024$. At evaluation time, images are resized to 256 pixels on the shorter side and center-cropped to $224 \times 224$. We report top-1 accuracy on the ImageNet-1K validation set.

\subsection{Semantic Segmentation}

For semantic segmentation, we follow the protocol of~\citet{kerssies2024benchmark}. We evaluate ViT-S/16 and ViT-B/16 on ADE20K and Cityscapes by attaching a linear per-patch prediction head to the ViT encoder and fine-tuning the full model end-to-end. For pretrained checkpoints, only the encoder is initialized from the self-supervised weights; the segmentation head is initialized randomly.

We use AdamW~\cite{loshchilov2018decoupled} with polynomial learning-rate decay with power $0.9$. The segmentation head uses learning rate $10^{-4}$. For pretrained models, the encoder learning rate is scaled by $0.5$, while randomly initialized models use the same learning rate for encoder and head. We train for 40k steps on ADE20K and 20k steps on Cityscapes with batch size 16 and mixed precision.

Training augmentations include random horizontal flipping, color jitter, scale jitter in $[0.5, 2.0]$, padding when needed, and random cropping. Crop sizes are $512 \times 512$ for ADE20K and $1024 \times 1024$ for Cityscapes. Positional embeddings are interpolated to the target patch grid. At evaluation time, we use sliding-window inference with the same crop sizes and average logits in overlapping regions. We report validation mIoU.
\subsection{Depth estimation}
For depth estimation, we follow the evaluation protocol used by \citet{yang2025pixio}, but perform full-finetuning instead of evaluating frozen encoders. 
Hence, we set the encoder learning rate to half that of the DPT~\cite{ranftl2021vision} depth head.
The prediction head takes normalized patch tokens and CLS tokens from four intermediate encoder layers as input, concatenated along the channel dimension.
We set the minimum depth to 0.001 for both datasets and the maximum depth to 10 and 80 for NYUv2 and KITTI, respectively.
We train using the scale-invariant logarithmic loss (SiLog loss) for 60 epochs with a batch size of 32, a learning rate of $1\times10^{-4}$ for the depth head, and a weight decay of $0.01$.
We use linear warmup for the learning rate in the first $10\%$ of training. 
During training, we use a crop size of $256\times256$. During validation, we employ sliding-window inference with a stride of $\frac{2}{3}$ of the crop size, corresponding to 170 pixels, and average predictions in overlapping regions. For depth estimation, the reported metrics are mean $\pm$ standard deviation over two fine-tuning runs with different seeds.

\section{Compute Resources}
\label{sec:appendix_compute}

We report approximate compute requirements in Table~\ref{tab:compute_wtp}. All runs were performed on NVIDIA H100 GPUs with mixed-precision training. Wall-clock times for pretraining exclude downstream fine-tuning. The full research project required additional compute for preliminary experiments, hyperparameter sweeps, and failed runs that are not included in the table.

For \ours{}, a single H100 is sufficient, but we typically used $2\times$H100 for \wtp{} and $4\times$H100 for longer streams to reduce turnaround time. On \wtp{}, \ours{} takes approximately 7 hours with ViT-S/16 and 8 hours with ViT-B/16 on $2\times$H100, and scales roughly with pretraining stream duration. MoCo v3 and DINO on \wtp{} take approximately 18 and 25 hours, respectively, on $2\times$H100. For these runs, \text{DataDrop} is enabled with $B=2048$, retaining $25\%$ of samples.

Downstream full fine-tuning on ImageNet-1K takes approximately 4 hours for ViT-S/16 and 5 hours for ViT-B/16 on $4\times$H100. Semantic segmentation fine-tuning takes approximately 2 hours for ViT-S/16 and 2.5 hours for ViT-B/16 on a single H100. Depth fine-tuning takes approximately 3 hours on KITTI and 1.5 hours on NYUv2; reported depth metrics are mean $\pm$ standard deviation over two fine-tuning runs with different seeds.

\begin{table}[h!]
    \centering
    \footnotesize
    \caption{
    Approximate compute requirements. Pretraining times are reported for \wtp{} and exclude downstream fine-tuning.
    }
    \label{tab:compute_wtp}
    \setlength{\tabcolsep}{4.5pt}
    \begin{tabular}{llcc}
        \toprule
        Experiment & Backbone & GPUs & Wall-clock time \\
        \midrule
        \ours{} pretraining on \wtp{} & ViT-S/16 & $2\times$H100 & $\sim$7h \\
        \ours{} pretraining on \wtp{} & ViT-B/16 & $2\times$H100 & $\sim$8h \\
        MoCo v3 pretraining on \wtp{} & ViT-S/16 & $2\times$H100 & $\sim$18h \\
        DINO pretraining on \wtp{} & ViT-S/16 & $2\times$H100 & $\sim$25h \\
        \midrule
        ImageNet-1K fine-tuning & ViT-S/16 & $4\times$H100 & $\sim$4h \\
        ImageNet-1K fine-tuning & ViT-B/16 & $4\times$H100 & $\sim$5h \\
        Semantic segmentation fine-tuning & ViT-S/16 & $1\times$H100 & $\sim$2h \\
        Semantic segmentation fine-tuning & ViT-B/16 & $1\times$H100 & $\sim$2.5h \\
        KITTI depth fine-tuning & ViT-S/B & $1\times$H100 & $\sim$3h \\
        NYUv2 depth fine-tuning & ViT-S/B & $1\times$H100 & $\sim$1.5h \\
        \bottomrule
    \end{tabular}
\end{table}

\end{document}